%% file: mainV2.tex
\RequirePackage[2020-02-02]{latexrelease}
\documentclass{clv3}

\usepackage{xcolor}

\usepackage{arabtex}
\usepackage{utf8}
\setcode{utf8}

\usepackage{multirow}
\usepackage{multicol}
\usepackage{booktabs}

\usepackage{graphicx}
\usepackage[skip=0.333\baselineskip]{caption}
\usepackage{subcaption}

\begin{document}
\issue{1}{1}{2016}

\runningtitle{Dotless Representation of Arabic Text}

\runningauthor{Al-Shaibani and Ahmad}

\pageonefooter{Action editor: \{action editor name\}. Submission received: DD Month YYYY; revised version received: DD Month YYYY; accepted for publication: DD Month YYYY.}

\title{Dotless Representation of Arabic Text: Analysis and Modeling}



\author{Maged S. Al-Shaibani\thanks{Information \& Computer Science Department, KFUPM, E-mail: g201381710@kfupm.edu.sa.}}
\affil{King Fahd University of Petroleum and Minerals}

\author{Irfan Ahmad \thanks{Information \& Computer Science Department, KFUPM, E-mail: irfan.ahmad@kfupm.edu.sa.}
\thanks{SDAIA--KFUPM Joint Research Center for AI, Dhahran 31261, Saudi Arabia.} 
}
\affil{King Fahd University of Petroleum and Minerals}




\maketitle

\begin{abstract}
This paper presents a novel dotless representation of Arabic text as an alternative to the standard Arabic text representation. We delve into its implications through comprehensive analysis across five diverse corpora and four different tokenization techniques. We explore the impact of dotless representation on the relationships between tokenization granularity and vocabulary size and compare them with standard text representation. Moreover, we analyze the information density of dotless versus standard text using text entropy calculations. To delve deeper into the implications of the dotless representation, statistical and neural language models are constructed using the various text corpora and tokenization techniques. A comparative assessment is then made against language models developed using the standard Arabic text representation. This multifaceted analysis provides valuable insights into the potential advantages and challenges associated with the dotless representation. Last but not the least, utilizing parallel corpora, we draw comparisons between the text analysis of Arabic and English to gain further insights. Our findings shed light on the potential benefits of dotless representation for various NLP tasks, paving the way for further exploration for Arabic natural language processing.
\end{abstract}


\section{Introduction}

Arabic serves as the official language across 22 nations within the Arab world, wielding a native speaker base exceeding 400 million individuals \cite{guellil2021arabic}. Moreover, it functions as a secondary language for numerous non-Arabic adherents within the Muslim community. Recognized as one of the United Nation's six official languages \cite{boudad2018sentiment}, Arabic occupies a significant linguistic sphere. Notably, it has ascended to become the fourth most prevalent language on the internet, experiencing rapid growth in user engagement between 2013 and 2018 \cite{boudad2018sentiment}.

The language itself exhibits a tripartite classification. Classical Arabic (CA) represents the linguistic embodiment within sacred texts, tightly interwoven with Islamic cultural and religious heritage, as well as ancient Arabic literature. Modern Standard Arabic (MSA) constitutes the formal register employed in official documentation and news dissemination. Finally, Dialectical or Colloquial Arabic (DA) encompasses a multitude of regionally specific vernaculars derived from MSA, utilized extensively in day-to-day interpersonal communication.

Arabic, akin to various Semitic languages, is distinguished for its dense morphology, characterized by a non-concatenative structure. In this morphological framework, words are crafted by interweaving vowel letters amidst the root consonants, coupled with the incorporation of prefixes and affixes. This intricate morphology poses formidable challenges for prevailing natural language processing (NLP) tools. For instance, the expansion of the language's lexicon surpasses that of languages such as English. Research by \citet{alotaiby2009processing} underscores this, revealing that the vocabulary size within an Arabic corpus is twice that of an English corpus from a commensurate domain, despite containing a roughly equivalent number of tokens.

\begin{figure}
    \centering
    \includegraphics[scale=0.45]{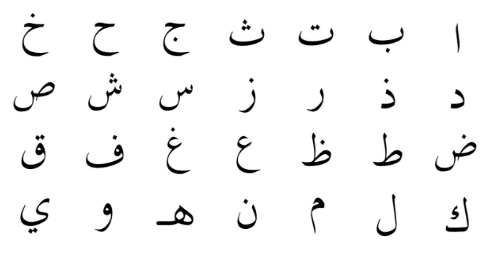}
    \caption{Characters in the Arabic script.}
    \label{fig:dotted_letters}
\end{figure}

The Arabic script, a cursive writing system, follows a right-to-left orientation. Each letter within this script exhibits diverse glyphs contingent upon its placement within a word. Comprising a total of 28 letters, 25 of these represent consonants. Notably, nearly half of these letters share fundamental base glyphs, known as \textit{rasm}, but are differentiated by the presence of dots either atop, within, or below their \textit{rasm}, a visual distinction elucidated in Figure \ref{fig:dotted_letters}. Additionally, the script incorporates 8 supplementary symbols known as diacritics. These diacritics serve the purpose of offering phonetic guidance, thereby mitigating potential ambiguity.

\begin{figure}
    \centering
    \includegraphics[scale=0.17]{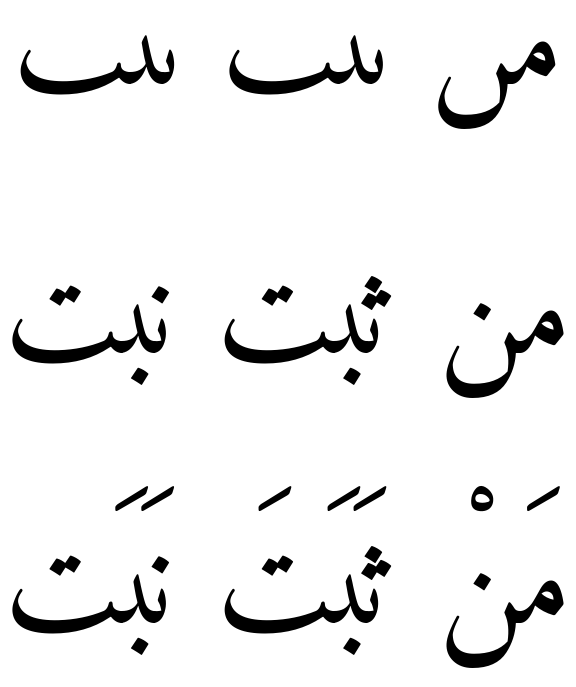}
    \caption{Arabic text with and without dots and diacritics}
    \label{fig:text_types}
\end{figure}

During the initial stages of the language writing system development, the utilization of dots lacked prominence and formalization \cite{abdo2020Thawabit}. This stemmed from the limited popularity of writing practices during that era, coupled with the relative ease with which native speakers could disambiguate dotless text within its contextual framework. The inception of standardized dot usage can be attributed to Abu Al-Aswad Al-Dualli, a renowned Arabic scholar, who introduced dots in the sacred text of the Holy Quran. These dots functioned akin to diacritics, with particular emphasis placed on the final letter in a word, representing the word's grammatical function within a sentence. However, this proposal was subsequently supplanted by the utilization of dots to differentiate letters sharing identical \textit{rasm} and diacritics to emulate phonetic assistance.

The primary motivation behind this endeavor was to facilitate Arabic comprehension for non-native speakers 
thereby safeguarding against foreign influences on subsequent generations of native speakers. Figure \ref{fig:text_types} showcases an Arabic sentence depicted in three distinct formats. The first is the dotless sentence, followed by the sentence with dots, and finally, the sentence incorporating both dots and diacritics. Notably, the second word illustrates the potential for multiple meanings, a disparity resolved through the use of dots.

Considerable attention has been devoted to the automatic diacritization of Arabic text, as evidenced by extensive research efforts documented in works such as \cite{diab2007arabic,abandah2015automatic,metwally2016multi,mubarak2019highly}. This emphasis is primarily due to the potential utility of diacritization in facilitating downstream tasks like machine translation \cite{diab2007arabic}, word sense disambiguation and part-of-speech tagging \cite{darwish2017arabic}. Comprehensive surveys conducted by \citet{hamed2017survey} and \citet{azmi2015survey} aim to delineate the advancements achieved in this domain.

Conversely, in the realm of Optical Character Recognition (OCR), the treatment of dotless text has garnered attention. Certain studies have explored techniques focused on the separation of dots from their respective \textit{rasms} during the OCR process \cite{ahmad2015multi,ahmad2019handwritten}. However, the utilization of dotless text in natural language processing tasks, such as sentiment analysis or language modeling, remains, to the best of our knowledge, an unexplored avenue. This presents a promising area warranting further investigation and potential exploration.

Before delving into an experimental analysis to evaluate the applicability and utility of dotless Arabic text as a representation for NLP tasks, conducting a preliminary statistical analysis would offer valuable insights. This statistical examination encompasses token counts across multiple levels of granularity, namely words, subwords, and characters. 

Additionally, employing quantitative analytical tools such as Zipf's law \cite{zipf1949principie} would be beneficial. Zipf's law elucidates an exponential relationship between the frequency of vocabulary items $v_i$ and their respective rank. Extensively studied across numerous languages \cite{yu2018zipf}, literary styles, and varying degrees of morphological abstraction \cite{moreno2016large}, Zipf's law has demonstrated applicability across diverse linguistic units, encompassing phrases, pragmatics \cite{sicilia2002extension}, spoken dialogues \cite{linders2022zipf}, and even in domains seemingly unrelated to natural languages, including music \cite{serra2012measuring}, users' telephone call patterns \cite{newman2005power}, programming languages \cite{sano2012zipf}, constructed languages like Esperanto \cite{manaris2006investigating}, and even in chess openings \cite{blasius2009zipf}.

Furthermore, Heap's law \cite{heaps1978information} serves as another valuable analytical tool. This law explores the growth of vocabulary within a given corpus relative to its size. According to Heap's law, the expansion of vocabulary exhibits exponential growth proportional to the corpus size, characterized by an exponential parameter $\beta$ typically less than 1. This implies that the growth rate of vocabulary lags behind the rate of corpus size expansion, elucidating a slower-than-linear increase in vocabulary size. Employing these statistical tools can provide foundational insights essential for subsequent experimental analyses.

Employing dotless Arabic text presents a promising avenue to address various challenges encountered in Arabic natural language processing (ANLP). The inherent density of Arabic morphology often results in a considerably expansive language vocabulary. However, dotless text adoption could aid in mitigating this issue by consolidating many dotted words into a singular dotless homographic word. This consolidation contributes to reducing the overall vocabulary size, offering a potential solution within ANLP.

Moreover, this line of inquiry holds significant relevance in the realm of automatic recognition of ancient parchments where the script used was predominantly dotless. By leveraging dotless text methodologies, researchers and practitioners can enhance their capabilities in deciphering and analyzing historical documents etched in dotless Arabic script.

Furthermore, this study's findings can be extended to augmenting processing tools for social media platforms, especially amid a recent trend encouraging users to employ dotless text instead of dotted text in posts that contravene platform regulations \cite{rom2021supporting,alimardani2021digital}. These studies have highlighted that despite the absence of dots, native speakers retain a substantial ability to recognize and interpret dotless text, underscoring the resilience and adaptability of native speakers within this context. Incorporating dotless text methodologies could serve as a means to address platform regulation breaches while acknowledging the inherent readability of dotless text by native speakers.

Our research interest is deeply rooted in the historical context wherein ancient Arabs adeptly interpreted dotless text, motivating us to undertake a comprehensive analysis encompassing both dotted and dotless textual dimensions. Our primary objective is to assess the efficacy of context-aware deep learning models specifically when applied to dotless text, contrasting their performance with the dotted variant. To explore this, we propose leveraging language modeling as a fundamental task within NLP, wherein we meticulously analyze the perplexity of text generated from both dotted and dotless formats.

We also hypothesize that through harnessing sufficiently extensive contextual information, these sophisticated algorithms hold the potential to achieve a level of proficiency in disambiguating dotless text that rivals human-level comprehension. This pursuit seeks to elucidate the capacity of modern computational models to grasp and interpret dotless Arabic text, drawing parallels with the interpretive capabilities of historical Arab scholars.

To summarize the contribution of this paper, it presents:
\begin{itemize}

    \item A novel method to represent Arabic text by removing dots from the Arabic letters.
  
    \item A comprehensive analysis of Arabic text rigorously comparing dotted and dotless text conducted on five different datasets, varying in size and domains. The analysis encompasses multiple tokenization levels, specifically character, word, and subword tokenizations. Additionally, this study supplements its findings by conducting a comparative analysis between Arabic dotted and dotless text and English text, enhancing the depth and breadth of the research insights.
    
    \item An extensive examination and comparison of dotted and dotless text conducted with two primary language modeling techniques – statistical n-grams and neural language models – utilizing corpora sourced from diverse domains. This analysis encompasses various tokenization schemes. To the best of our knowledge, no existing literature comprehensively covers this specific analysis, making it a valuable contribution to the field of language modeling techniques.

\end{itemize}


The remaining sections of this paper are organized as follows: Section \ref{sec:related_word} delves into the advancements and developments pertinent to this topic in the existing literature. Section \ref{sec:corpora} provides an overview of the corpora utilized in this study, detailing the preprocessing steps undertaken and elucidating the employed tokenization methods. In Section \ref{sec:text_analysis}, we present the various text analyses conducted to compare dotted and dotless text. Section \ref{sec:lm} outlines our experiments involving language modeling. Section \ref{sec:compare_with_en} offers a comparative analysis between dotted, dotless text, and English text to maintain a sense of broad applicability and generalization. Finally, Section \ref{sec:conclusion} concludes this study, encapsulating the findings and envisaging future directions for research.

\section{Related Work}
\label{sec:related_word}


As far as our research indicates, this is the first work to analyze dotless Arabic text in the domain of natural language processing tasks. However, the utilization of dotless text in social media to circumvent content filtering algorithms has attracted recent attention \cite{alimardani2021digital}. In response to this trend, \citet{rom2021supporting} delved into methodologies to accommodate undotted text within the prevailing settings of language models trained predominantly on dotted text, particularly in the context of social media platforms. Their exploration yielded two primary approaches. The first approach involved reconfiguring the tokenizer either by removing dots from its tokens or expanding it with undotted variants of these tokens. These undotted versions were then mapped to the same identifier as their dotted counterparts, effectively enabling the language model to accept both dotted and dotless text. The second approach proposed a dotting algorithm aimed at restoring dots to the undotted text. Through this method, they demonstrated that the restored dotting approach produced results closely aligned with the original experiments conducted on dotted text when evaluated across various downstream tasks.

\citet{alhathloul2022automatic} presented a deep learning-based approach aimed at storing dots for a given dotless text. Employing an extensive experimentation strategy across various datasets, they conducted an in-depth error analysis to assess the efficacy of their proposed method. Their approach relied on a Gated Recurrent Networks (GRUs) based sequence-tagging model, designed to generate dotted text as an output from dotless text input. In order to facilitate a more equitable comparison and evaluation, the researchers introduced a new metric called "dottization error rate." Across their experimentation involving four distinct datasets, their method exhibited promising results, achieving character error rates ranging between 2\% and 5.5\%, along with dottization error rates spanning from 4.2\% to 11\%.

Zipf's and Heap's laws serve as pivotal statistical tools in linguistics, offering profound insights when applied to language corpora, extensively examined in languages like English, Chinese, Norwegian, and Old German \cite{li1992random,moreno2016large,sicilia2003extension,welsh1988codes}. However, their application to Arabic remains relatively limited, potentially influenced by Arabic's perception as a low-resource language \cite{magueresse2020low}. Within Arabic language studies, two primary research trajectories emerge. Firstly, analytical comparisons of Arabic with other languages have been explored \cite{alotaiby2009processing,alotaiby2014arabic,al1998study}. Secondly, these analyses have been utilized to assess the quality of proposed corpora \cite{alarifi2012estimating,almeman2013automatic,selab2015building,khreisat2009machine}. Notably, much of this research has focused on evaluating extensive corpora within a singular domain, predominantly newswire. Consequently, a notable research gap exists in conducting a comprehensive analysis encompassing diverse tokenization levels, linguistic units, and datasets spanning various domains within Arabic language studies.

Language models constitute mathematical models designed to predict the most probable word within a sequence of preceding words, representing a critical component in numerous natural language processing endeavors. Their utility extends across various tasks, including generative applications like machine translation \cite{diab2007arabic}, optical character recognition enhancement \cite{smith2011limits}, and the augmentation of automatic speech recognition systems \cite{abushariah2010natural}. Leveraging their capacity to encapsulate implicit linguistic information, neural language models find applications in sequence labeling tasks such as part-of-speech tagging and named entity recognition through a transfer learning style. Within these contexts, language models play a pivotal role in generating high-quality embeddings to enhance subsequent classification processes.


The evolution of language modeling boasts a substantial developmental history. Initially, these models sought to learn the joint probabilities of sequences within expansive training corpora. Employing the Markov principle, these sequences were pruned to encompass orders 2, 3, 4, or higher $n$ tokens, commonly known as n-grams, aiming to achieve an acceptable level of generalization. However, this approach grappled with an inherent challenge to dealing with Out-Of-Vocabulary (OOV) tokens, wherein the model encountered difficulty computing the probability of words absent from the training corpus. To address this hurdle, various smoothing techniques emerged, including Laplace (additive) smoothing, Katz smoothing \cite{katz1987estimation}, and Kneser-Ney smoothing \cite{kneser1995improved}. Subsequently, specialized toolkits emerged as standardized solutions for constructing these language models using sufficiently extensive corpora. Prominent examples of such toolkits include KenLM \cite{heafield2011kenlm} and SriLM \cite{stolcke2002srilm}. These toolkits have streamlined the development process by providing efficient means to build robust language models, marking significant strides in the field of language modeling.

Neural language models surpassed statistical counterparts \cite{bengio2000neural}, with Recurrent Neural Networks (RNNs) initially dominating sequence modeling. However, they grappled with issues like vanishing or exploding gradients, particularly evident in lengthy sequences. To address this, Long Short-Term Memory (LSTM) architectures were introduced \cite{hochreiter1997long}, albeit slower and more challenging to train. Gated Recurrent Units (GRUs) \cite{chung2014empirical} emerged as a compromise between Vanilla RNNs and LSTMs. Recently, transformer-based models \cite{vaswani2017attention} sparked a revolution in Natural Language Processing (NLP), overcoming the long recurrence limitation of previous recurrent networks. Pretrained models like BERT \cite{devlin2018bert}, Wav2Vec2 \cite{baevski2020wav2vec}, and GPT-3 \cite{brown2020language} attained state-of-the-art performance across various NLP tasks using this architecture. Nonetheless, the attention mechanism, intrinsic to transformers, exhibited limitations in mastering simpler tasks easily tackled by earlier models \cite{dehghani2018universal,chernyavskiy2021transformers}.



\section{Text Corpora, preprocessing and Tokenization}
\label{sec:corpora}

In this section, we introduce the corpora utilized in this research, outlining the preprocessing steps employed in their preparation. Additionally, an overview of the tokenization methods applied is presented for comprehensive understanding and context within this study.

\subsection{Text Corpora}

We selected corpora from different domains with different sizes to evaluate dotless text as an alternative representation for Arabic NLP. The following are the selected corpora:

\begin{itemize}
    \item \textbf{Quran} \cite{Aloufi2019-lr}: a dataset compiled from Quran verses. In Quran, most chapters, called Surah, start with a special sentence, called Basmala. In order not to confuse the training with this repetitive sentence, we removed it from the beginning of each chapter except the first one. The total number of samples is 6236.
    
    \item \textbf{Sanadset dataset} \cite{mghari2022sanadset}: This dataset is a collection of Sacred narrates by the prophet Muhammed called Hadeeth. Each Hadeeth contains a Matn which is the sacred text and a Sanad which is the chain of narrators. We selected only the Matn. This dataset is interesting because its text is coherent with a relatively small vocabulary size compared to its size.
    
    \item \textbf{Ashaar (Poems)} \cite{alyafeai-al-shaibani-2020-arbml}: This is a poetry dataset collected from various online sources. Samples extracted from this dataset comprises a set of 6 verses. Throughout this research, we will refer to this dataset as "Ashaar" or "Poems" dataset. 
    
    \item \textbf{Newswire} dataset \cite{el20161}: This dataset is a massive collection of news collected from various news wires. We only selected one newspaper called Alittihad. 
    
    \item \textbf{Wikipedia} \cite{wikidump}: a dataset collected from Wikipedia dumps of Arabic articles up to 20-10-2022. 
    
\end{itemize}

Table \ref{tab:datastes_stats} presents detailed statistics with respect to dotted and dotless tokens for each dataset. In this table, N represents the running text, i.e. all words, V is the unique vocabulary, and V` is the unique dotless vocabulary. As illustrated in the table, there is a noticeable reduction of the dotless vocabulary as compared to dotted. It can also be noticed that the Quran dataset is the smallest dataset compared to others. For Sanadset dataset, it contains very coherent text with many repetitive phrases. This can be deduced from the ratio of its unique vocabulary to its running text. This measure highlights the abundant richness of the poetry dataset, which aligns with our expectations, given the inherent creativity associated with poetry. Furthermore, classical poetry adheres to strict metrical and rhyming constraints, which compel poets to seek words that harmonize with the poem's rhyme scheme. Regarding news and Wikipedia datasets, they are huge corpora capturing various aspects and structures of the language. However, the news dataset has a much narrower domain as compared to Wikipedia.

From the above statistics, it is clear that the datasets we chose are of different sizes and can be grouped into three categories. From the size of the running text N, the Quran text is a significantly small corpus, Sanadset and Ashaar are medium-sized corpora, Wikipedia and news are larger corpora.

\input{tables/datasets_statistics}


\subsection{Text pre-processing and undotting}

In this subsection, we present the procedure we followed to pre-process and undot text. For pre-processing, we removed any non-Arabic characters including numeral characters and punctuation symbols. Diacritics are also omitted. Further, as the letter Hamza can appear adjoined with other characters, usually vowels, we removed it keeping only the adjoined letter. However, if it comes alone, it is kept unchanged.

For undotting, table \ref{tab:undotting_mapping} illustrates Arabic letters and their dotless mapped letter. Due to the cursive nature of the Arabic script, we tried to mimic the same writing shape for a few of the letters to its closest dotless letter if it comes either at the beginning or the middle of the word. For the letters \RL{ن} and \RL{ي}, if they come either at the beginning or the middle of the word, they are mapped to the dotless version of \RL{ب}. Similarly, for the letter, \RL{ق}, if it comes at the beginning or middle of the word, it is mapped to the dotless version of \RL{ف}.

\input{tables/undotted_mapping_table}



\subsection{Tokenization}

Tokenization stands as a crucial process essential for converting text into integer identifiers, serving as inputs for deep learning systems. This transformation involves segmenting text into various levels of granularity, including words, characters, or in-between units known as subwords, achieved through a meticulous segmentation procedure. Notably, subword tokenization emerges as an intriguing approach due to its adaptability, being either language-dependent or language-agnostic. This method integrates benefits from different ends of the spectrum, presenting a versatile and robust means to represent input text. The proliferation of large language models such as BERT \cite{devlin2018bert} and GPT-3 \cite{brown2020language} significantly contributed to the popularity of subword tokenization. Data-driven approaches like WordPiece \cite{song2020fast} and SentencePiece \cite{kudo2018sentencepiece} exemplify such tokenization methodologies, underscoring their significance in modern language modeling techniques.

In this research, we studied the behavior of dotless text compared to dotted text in the context of these levels of granularity. We used word tokenization, character tokenization, morphology-based, and language-specific subword tokenizations. These tokenizations are described as follows.

\begin{itemize}
    \item \textbf{Word tokenization}: This tokenization splits text into words.
    \item \textbf{Farasa Morphological tokenization}: This tokenization splits text into morphemes. That is roots and affixes. The tool used to deliver this tokenization is called farasa \cite{abdelali2016farasa}, hence the name. We used this tokenization to represent morphological splits of the Arabic text.
    \item \textbf{Disjoint-Letters tokenization}: This tokenization is a language dependent tokenization introduced by \cite{alyafeai2022evaluating}. Although the Arabic script is cursive, some of its letters are written disconnected from their subsequent letter in the word. Based on this property, this tokenization splits the text into subwords according to this property of the letters. That is, each subword in this tokenization is a sequence of fully connected letters. 
    \item \textbf{Character tokenization}: This tokenization splits text into character. 
\end{itemize}

\begin{figure}
    \centering
    \includegraphics[scale=0.4]{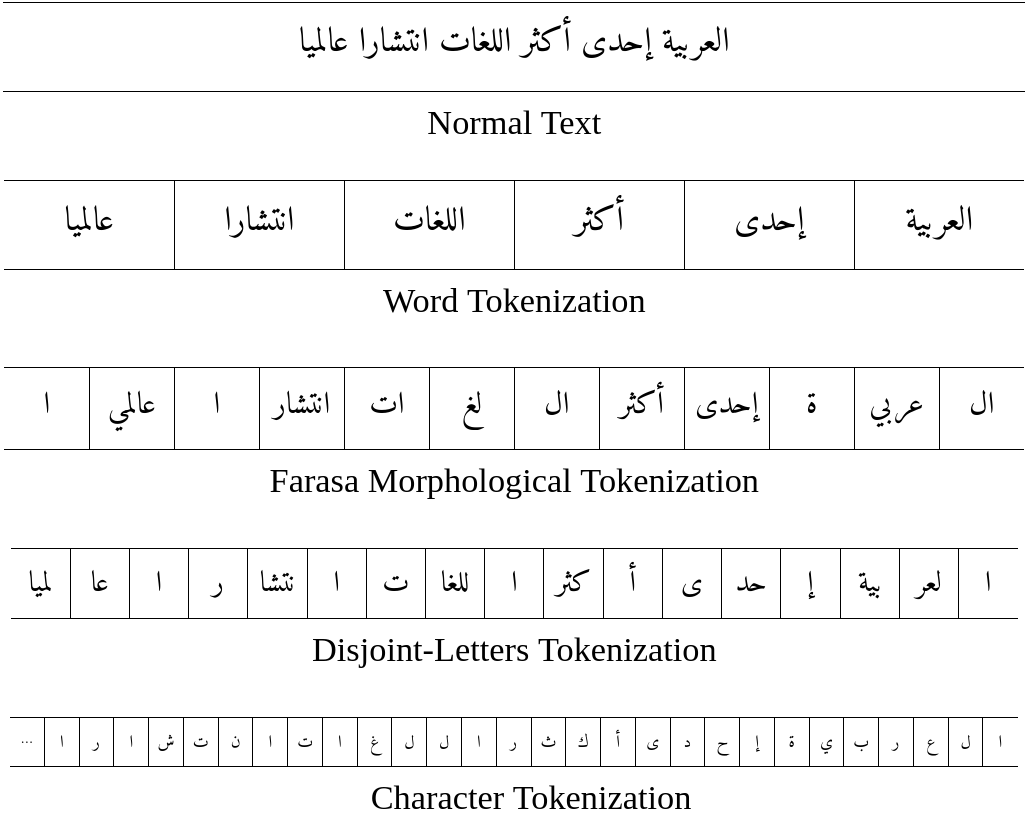}
    \caption{An Arabic sentence tokenized with different tokenizations}
    \label{fig:tokenizations}
\end{figure}

Figure \ref{fig:tokenizations} shows an example of an Arabic sentence being tokenized with the aforementioned tokenizations.

\citep{alyafeai2022evaluating} conducted an in-depth comparison of various types of tokenization for Arabic where this tokenization was first introduced for NLP tasks. To perform the tokenization task, we extended their tokenizers package, named tkseem \footnote{https://github.com/ARBML/tkseem}, to fit our needs while utilizing their implemented functionalities.


\section{Text Analysis}
\label{sec:text_analysis}

This section delineates the experimental framework established to compare dotless text with its dotted counterpart. Initially, it provides a comprehensive exposition of detailed vocabulary statistics spanning diverse tokenization levels. Subsequently, it delves into an examination of dotless and dotted text within the purview of Zipf's and Heap's laws, offering insights into their respective applicability and manifestations. Furthermore, it encompasses a subsection dedicated to elucidating the outcomes derived from conducting text entropy analysis over both dotted and dotless text, providing a nuanced understanding of their respective linguistic properties and information content.

\subsection{Vocabulary Statistics}

As this work presents a new representation for writing Arabic text based on the removal of the dots, it is a valuable endeavor to discuss dotless vocabulary as compared to dotted ones on all our datasets and across multiple tokenizers. Tables \ref{tab:words_stats}, \ref{tab:farasa_stats}, and \ref{tab:disjoint_stats} present such statistics for words, farasa, and disjoint tokenizations, respectively. It also presents the average tokens' length for each dataset. In these tables, V` represents dotless vocabulary, S is the average vocabulary length for the dotted text, and S` represents average vocabulary' length for dotless text. 

\input{tables/tokens_stats/words}
\input{tables/tokens_stats/farasa}
\input{tables/tokens_stats/disjoint}

Table \ref{tab:words_stats} serves to elucidate the vocabulary statistics pertaining to words tokenization across various datasets. This table reveals notable insights, particularly in gauging vocabulary richness, denoted by the size of V in comparison to N. In smaller datasets like Quran, the value of this metric appears notably high, contrasting with sufficiently large datasets where this value may even fall below 1. This reduction, expounded by Zipf's and Heap's laws, illustrates that within expansive datasets, a small subset of V accounts for a substantial portion of the running text N. Delving deeper, disparities in vocabulary richness among datasets become apparent, such as Ashaar being richer than Sanadset. This discrepancy arises from the inherent constraints imposed by the creativity, classical metric, and rhythmic structure in classical Arabic poetry, necessitating a broader vocabulary to adhere to these constraints. Conversely, Sanadset exhibits higher consistency and lesser diversity in topics, resulting in a less rich vocabulary. Furthermore, in the comparison between Wikipedia and news datasets, the former displays higher richness owing to its diverse domains and topic variety, whereas the latter, sourced from fewer domains with more consistent article structures, manifests a lower richness. These observations underscore the influence of content diversity and structural consistency within datasets on their respective vocabulary richness, offering valuable insights into their distinct linguistic characteristics.



The comparison of V with respect to N for subword tokenizations is detailed in Table \ref{tab:farasa_stats} for farasa tokenization and Table \ref{tab:disjoint_stats} for disjoint tokenization. Evident from both tables is the finer granularity of disjoint tokenization compared to farasa, indicating a higher prevalence of disjoint letters within the language vocabulary in contrast to morphologically segmented components. Moreover, the discrepancy observed between V for farasa and disjoint tokenizations is influenced by two key factors: the extent of running text (N) and the inherent richness of the dataset. Notably, this discrepancy is more pronounced when datasets exhibit richness and consequently higher N values. This trend is particularly observable in datasets like Wikipedia and Ashaar within the dotted text domain, where the difference in V between farasa and disjoint tokenizations is notably elevated. This delineates the influence of dataset richness and the volume of running text on the variance between farasa and disjoint tokenizations, emphasizing their impact on the vocabulary distinctions observed across different tokenization schemes. 



A clear trend observable across all tokenizations from these tables is that S`—representing the average vocabulary length for dotless text—is generally marginally higher than S, denoting the average vocabulary length for dotted text. This recurrent pattern underscores the characteristic distinction between least and most frequent vocabulary items across various tokenizations and the subtle disparity in average vocabulary lengths between dotless and dotted text.

From these tables, it can be noticed that rich datasets characterized by richness tend to exhibit smaller V`/V ratios across almost all tokenizations. For example, the Wikipedia dataset showcases lower V`/V values for words and disjoint tokenization compared to the News dataset, and a comparable ratio in farasa tokenization, despite having a considerably larger running text N. This pattern is even more apparent in the poetry dataset, which boasts the smallest V`/V ratio owing to its richness in vocabulary. This observation leads to a hypothesis that, asymptotically, V` grows slower concurrently with N in the long run. In essence, it suggests that, generally, the least frequent vocabulary items tend to share more \textit{rasms} compared to the more frequent ones, leading to a proportional growth of dotless vocabulary as the volume of running text expands.

The observation that farasa's V`/V ratio is higher than Word's V`/V for datasets with high vocabulary as can be shown in table \ref{tab:farasa_stats} is of a particular interest. This discrepancy suggests that subwords segmented by farasa might not share as many \textit{rasms} when dots are removed, in comparison to words tokenization. This disparity in the V`/V ratio between farasa and words tokenizations highlights a potential difference in the underlying structure of segmented subwords, indicating that the removal of dots might affect these segmented subwords differently in terms of shared \textit{rasms}.

An other interesting finding is the consistently smaller V`/V ratio observed for disjoint tokenization compared to other tokenizations. This ratio is even smaller than the characters' V`/V ratio in the three largest datasets. This discrepancy suggests that these subwords segmented by disjoint tokenization are notably rich in dots and, consequently, exhibit a higher likelihood of sharing the same \textit{rasms} upon the removal of dots. This observation underscores the distinct characteristic of disjoint subwords, implying a stronger association or shared \textit{rasms} among these segments when dots are excluded.

Another final notable observation is the consistently lowest V`/V ratio observed in the aggregated dataset across all tokenizations. This observation implies that with larger datasets, a greater proportion of \textit{rasms} are being shared. This trend suggests that as the dataset size increases, there is a higher tendency for shared \textit{rasms} among vocabulary.

\begin{figure}[!ht]
    \centering
    \includegraphics[scale=0.35]{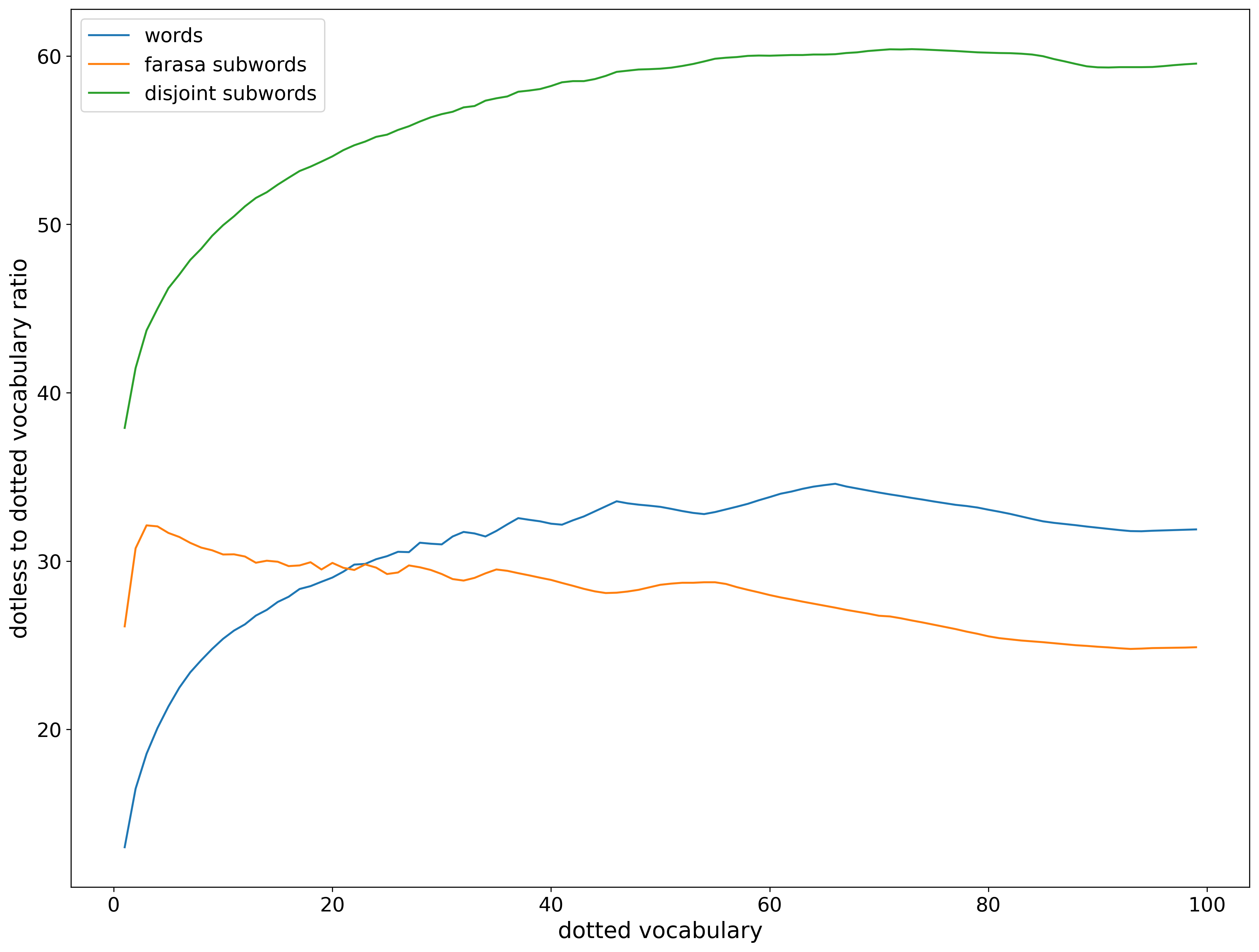}
    \caption{Dotless vocabulary reduction ratio on different tokenization levels}
    \label{fig:dotless_vocab_analysis}
\end{figure}

To explore the correlation between dotted and dotless vocabulary, we conducted an analysis on the top $i$ frequently occurring vocabularies, ranging from 1 to 100. Subsequently, we transformed these vocabularies into their dotless forms and calculated the ratio between dotless and dotted vocabulary counts. This analysis was conducted across aggregated datasets employing words, farasa, and disjoint tokenizations. The findings of this analysis are represented in \ref{fig:dotless_vocab_analysis}.

Upon examination of the plot, a noteworthy observation emerges regarding vocabulary reduction. The most significant reduction occurs within the disjoint tokenization scheme, where the ratio of dotless to dotted vocabulary size experiences the highest reduction, starting from approximately a 40\% decrease. Additionally, it becomes evident that the reduction in vocabulary through words tokenization initially starts at a lower rate but increases as the vocabulary expands, surpassing the reduction observed in the farasa tokenization.

Moreover, across all three tokenization types, the ratio exhibits fluctuations, displaying intermittent rises and falls, forming plateaus. As a result of this analysis, we hypothesize that while many common vocabularies share identical \textit{rasms}, this ratio stabilizes as the vocabulary expands, corroborating our previous observations.

Furthermore, it is observed that within the farasa tokenization, the most frequently occurring vocabularies, anticipated to represent affixes, exhibit the highest number of shared \textit{rasms} compared to other tokenizations, whereas its least common vocabularies display the lowest number of shared \textit{rasms}. This inference suggests that affixes are notably abundant in dots compared to other vocabularies in this specific tokenization scheme.

Additionally, the reduction pattern observed in disjoint tokenization closely resembles that of words but appears smoother. Notably, the decrease in the ratio for the least common vocabularies in words exhibits a slightly faster descent compared to those in disjoint tokenization. This discrepancy implies distinct structural characteristics within these tokenization schemes.

\subsection{Zipf's and Heap's laws}

We applied Zipf's law to examine the correlation between vocabulary and its frequency. This empirical law delineates a seemingly straightforward exponential relationship between the rank of vocabulary and its frequency when the vocabulary is arranged in descending order based on frequency, particularly within a considerably large corpus. Our objective in this analysis was to visually explore the frequency distributions of both dotted and undotted vocabularies across varying corpus sizes.

Zipf's law can be mathematically expressed as shown in equation \ref{eq:zipf}, where $r$ signifies the rank of the vocabulary, $F(r)$ represents its frequency, and $\alpha$ stands for a corpus-specific parameter typically approximating unity. The primary aim of our investigation was to scrutinize how this law manifests itself in the context of the frequency distributions of dotted and undotted vocabularies across diverse corpus sizes.

\begin{equation}
     F(r) \propto r^{-\alpha}
\label{eq:zipf}
\end{equation}

To fit a corpus with Zipf's law and determine the value of $\alpha$, we applied a log transformation to both sides of the equation, as illustrated in equation \ref{eq:zipf_log}. By setting the constant $C$ as the frequency of the most frequent vocabulary within each dataset, we aimed to prevent potential biases when fitting the regression line. This transformation led to the formulation of an equation that exhibits near-linearity on a log-log scale, allowing us to employ linear regression to estimate the slope ($\alpha$).

\begin{equation}
     \log{F(r)}  = -C\alpha \log{r} \Longrightarrow \alpha =  - \frac{\log{F(r)}}{C\log{r}} 
\label{eq:zipf_log}
\end{equation}


In investigating the expansion of vocabulary alongside the running text within our corpus, we employed Heap's law \cite{heaps1978information}. This law similarly demonstrates an exponential correlation between the size of the running text denoted by $n$ and the corresponding vocabulary size denoted by $V$, governed by an exponent parameter $\beta$ typically falling below 1. The essence of this relationship is articulated in equation \ref{eq:heap}.

\begin{equation}
     V \propto n^{\beta} \Longrightarrow V = kn^{\beta}
\label{eq:heap}
\end{equation}

Following the same procedure conducted for Zipf's law to estimate the parameters, we transform both sides of the equation using the $\log$ function as in equation \ref{eq:heap_log}. We fit this equation using linear regression to estimate both values $k$ and $\beta$.

\begin{equation}
     \log{V} = \log{k} + \beta \log{n}
\label{eq:heap_log}
\end{equation}

\begin{figure}
    \centering
    \includegraphics[scale=0.32]{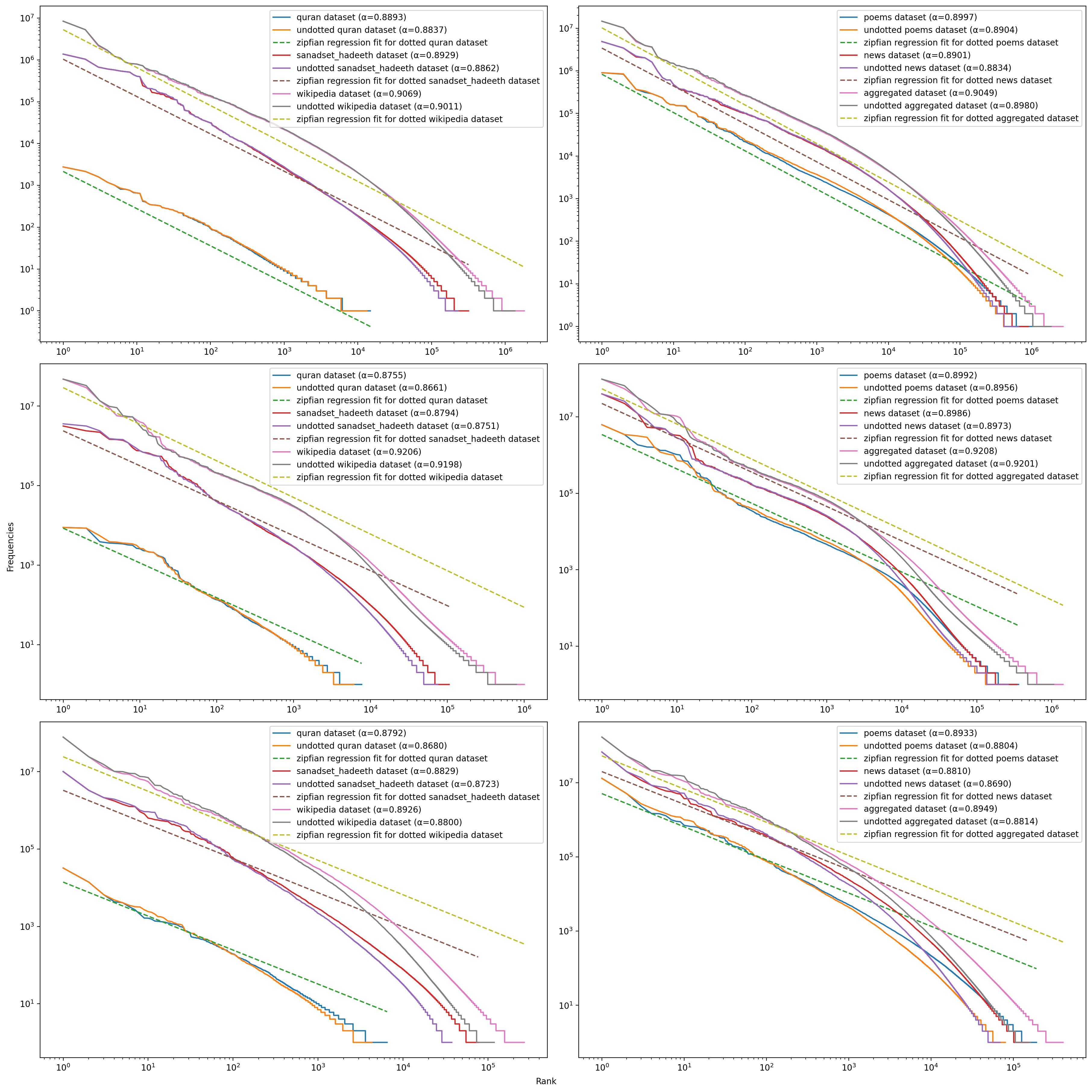}
    \caption{Zipf's Law plots for various datasets with different tokenizations}
    \label{fig:zipfs_plots}
\end{figure}

Figure \ref{fig:zipfs_plots} plots Zipf's law applied to our datasets. Each row represents a tokenization level. The first row is for word tokenization, the second row is for farasa tokenization, and the thrid row is for disjoint-letters tokenization. In each of these rows, datasets are split into two figures as it is difficult to plot all of them in one graph.

From an overarching perspective gathered from these graphs, it is evident that all tokenization methods adhere to Zipf's law. Nevertheless, the coarse-grained tokenizations exhibit a more pronounced adherence to Zipf's law, showcasing a stronger Zipfian distribution compared to other finer-grained tokenization approaches.

The depicted figure reveals a nearly identical pattern between dotless and dotted frequencies, with marginal distinctions apparent primarily within the most frequent and least frequent vocabularies. Notably, across various datasets, a consistent trend emerges where, with finer-grained tokenization methods, the dotless frequencies exhibit a slightly lower count. This discrepancy becomes particularly conspicuous in the disjoint plots. Remarkably, this observation remains consistent across all datasets, suggesting that this behavior is independent of the specific dataset characteristics.

Additionally, a distinct pattern emerges within the farasa vocabulary. Across all datasets, a consistent decline in the frequencies of vocabulary—both dotted and dotless—can be observed, occurring between what we can term as the most frequent and least frequent vocabularies. This drop in the frequencies of the least frequent vocabulary exhibits a consistent behavior, irrespective of the dataset or tokenization method employed. However, an intriguing observation arises concerning the decline in the frequencies of what we designate as the mid-frequent vocabulary. It appears that the most frequent vocabularies, presumed to represent affixes due to their high frequency resulting from their attachment to multiple roots, initiate this decline from the mid-frequent vocabulary and continue downward. This particular pattern seems to manifest across various datasets and other tokenization methods, suggesting a consistent relationship between these mid-level frequencies and the prevalence of affixes within the farasa vocabulary.

\begin{figure}
    \centering
    \includegraphics[scale=0.26]{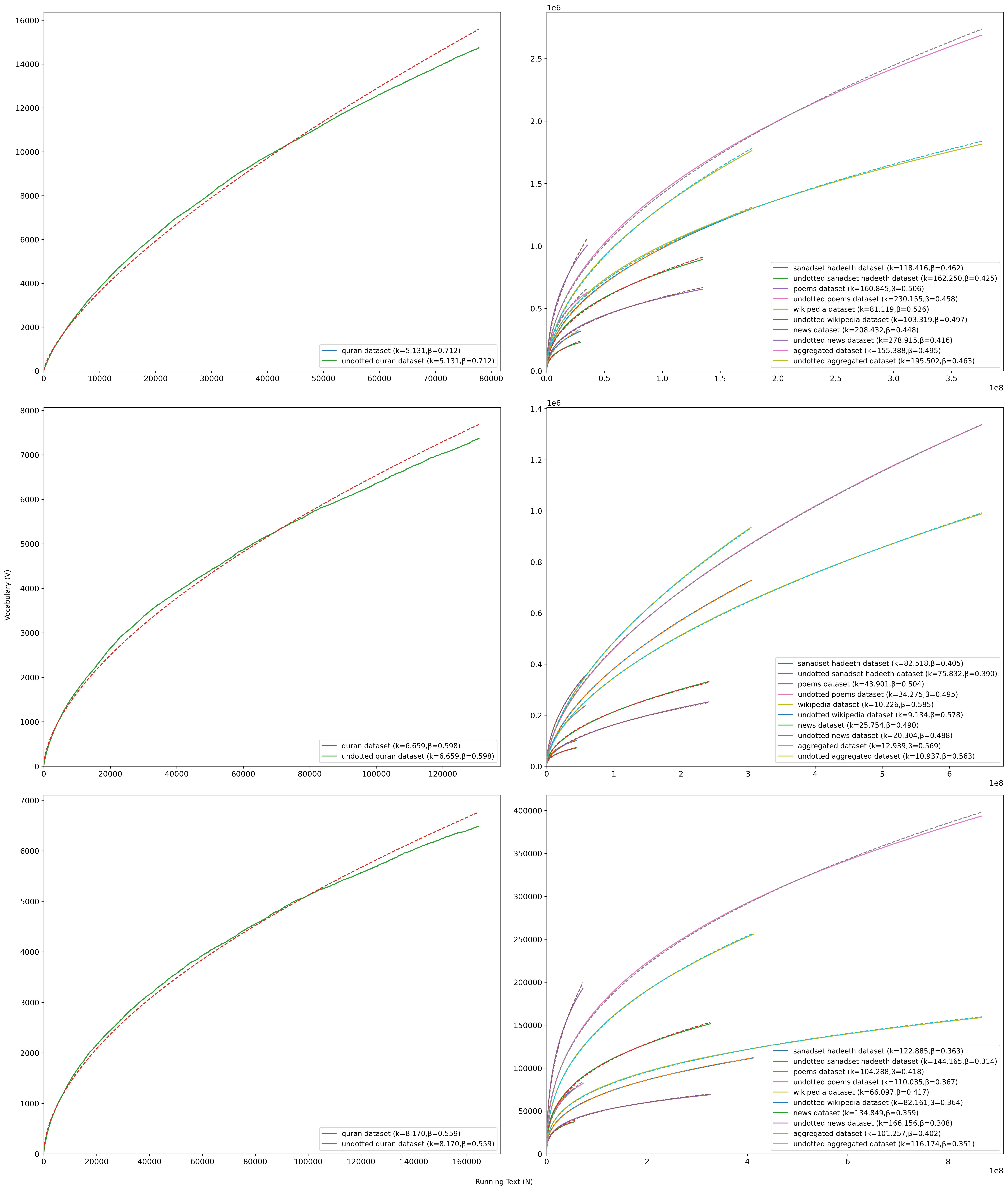}
    \caption{Heap's Law plots for various datasets with different tokenizations}
    \label{fig:heaps_plots}
\end{figure}

Figure \ref{fig:heaps_plots} illustrates Heap's law applied to various tokenizations across all datasets. Each row within the figure corresponds to a distinct tokenization type, while the first column depicts the plots for the Quran datasets, notably smaller in terms of running text compared to the other datasets. The subsequent column showcases the plots for the remaining datasets, exhibiting more substantial textual content in comparison.

The depicted plots demonstrate the conformity of both dotted and dotless vocabularies to Heap's law. Notably, the growth of dotless vocabulary consistently trails behind that of the dotted vocabulary across various tokenizations. However, a significant observation arises particularly in the case of disjoint tokenization where the growth rate appears notably slower. This phenomenon suggests a substantial sharing of \textit{rasms} among the dotted vocabularies within this specific tokenization scheme, contributing to a slower expansion of the vocabulary.

Another notable observation is the influence of dataset richness on the growth disparity between dotted and dotless vocabularies. Specifically, a more substantial gap between the growth rates of dotted and dotless vocabularies is noticeable in datasets with higher richness, such as Ashaar and Wikipedia. In contrast, this gap notably narrows in datasets characterized by lower richness, such as news and Sanadset. This divergence suggests a correlation between dataset diversity or size and the distinctiveness in growth rates between the dotted and dotless vocabularies.

\subsection{Text Entropy}

In the realm of information theory, entropy serves as a metric quantifying the degree of disorder within a given system. Its purpose lies in computing the number of bits necessary to transmit information, i.e., the information content of text in our context. In our investigation, we try to assess the extent of information loss resulting from the elimination of dots across various tokenization levels for all token strings, denoted as $T$, within our dataset. Equation \ref{eq:entropy} was employed to compute the entropy of tokens within the corpus, a dimension that has not been extensively explored in Arabic literature. Previous work by \citep{al1998study} delved into entropy calculations on Arabic text, reporting an entropy of 9.98 bits/words and 2.39 bits/letters. This study, considered a corpus comprising 3025 vocabularies, was conducted nearly two decades ago, predating recent advancements in computational resources. Surprisingly, no recent studies have delved into this particular aspect of Arabic text analysis. Table \ref{tab:entropy_stats} elucidates the entropy metrics for each tokenization scheme, where $H$ denotes dotted text entropy and $H`$ signifies dotless text entropy.

\begin{equation}
     H(t) = \sum_t^T-P(t)\log_2{P(t)} 
\label{eq:entropy}
\end{equation}

\input{tables/tokens_stats/entropy}

For character tokenization, we can see that the number of dotless characters drops by a margin of 12 letters with a ratio of 61.3\% of the dotted characters. However, we can also notice that the highest entropy for dotted characters is 4.3 and the lowest is 4.15 while the highest for the undotted are 3.9 and the lowest is 3.83. If we consider the aggregated dataset as a representation of the language, we can see that the reduction of entropy due to undotting is 0.4 with a ratio of 9.1\%. That is, the reduction in entropy is less compared to the reduction in the characters. From this, we can see that the undotted text still holds entropy close to the dotted one indicating only a little information being lost with this dotless representation compared to the reduction in the character set. This gives an indication that this type of text can be learned with a decent performance close to the dotted text. 

The upper bound of the entropy for characters can be calculated using equation \ref{eq:entropy} when we have a string of 1   character to be $-\sum_{i}^{31}P(i)*\log_2{P(i)}=\log_2{31}=4.95$ for dotted characters and $\log_2{19}=4.24$ for undotted ones. Based on this, if we consider the aggregated dataset as a representative set of the language, the language has a characters redundancy of $1-\frac{4.27}{4.95}=13.74\%$ for dotted and $1-\frac{3.87}{4.24}=8.76\%$ for undotted characters. The redundancy is less for undotted text because dots were helpful in resolving confusion. It can be concluded from this analysis that, dots, at most, decrease the text redundancy by 5\%. This also indicates that there are other sources of redundancy shared between dotted and undotted text.

Entropy, as discussed previously, measures the surprise in a sequence given a prior history of sequences. We can see that Ashaar has the highest entropy in word tokens. It is even higher than Wikipedia which is larger by magnitude in terms of running text N. This confirms our hypothesis about the vocabulary richness of Ashaar being due to creativity and the use of unusual phrases and sentences. We can also note that Sanadset has the lowest entropy. It is even lower than Quran dataset which is smaller by magnitude. This indicates that Sanadset samples share a lot of similarities and it is easily predictable.



As can be noticed in Table \ref{tab:entropy_stats}, the entropy results for subwords tokenizations are in consensus with words and characters. This is expected as this tokenization still preserves the datasets' characteristics. It is also interesting to see that the reduction in entropy is low compared to the reduction in vocabulary (V). This indicates that a little deeper model is required to capture the loss in entropy but with the advantage of much fewer vocabulary. This also confirms the choice of subwords tokenization as a standard method for language modeling in the current practice.

\section{Language Modeling}
\label{sec:lm}

Language modeling stands as a core area of research within NLP, focusing on generating text that mimics human language patterns. Its fundamental function involves learning the associations between tokens present in extensive text corpora. These models play a vital role in diverse downstream applications such as optical character recognition (OCR), speech-to-text (STT), and text grammar correction, among others.

Language models are constructed through either statistical methodologies or deep learning approaches. In statistical methods, text generation relies on determining the most likely token given a preceding sequence of tokens, computed from the corpus to establish conditional probabilities of tokens. These models, often termed n-gram models where $n$ denotes the sequence length of prior tokens considered, operate within a probabilistic framework. Conversely, neural language models employ neural networks to predict the most probable token, diverging from probabilistic strategies. Notably, statistical language models are typically shallower in structure compared to neural models, offering lower language comprehension abilities in their linguistic analyses.

Statistical language models encounter significant challenges in predicting unseen or infrequently used words due to their reliance on word probabilities computed from corpus counts. To mitigate this issue, smoothing techniques have been developed, including Laplace smoothing, discounting, interpolation, backoff, and notably, Kneser-Ney smoothing \cite{ney1994structuring}. Acknowledged as one of the most effective smoothing methods in literature \cite{zhang2014kneser}, Kneser-Ney smoothing operates by generating the subsequent token using Equation \ref{eq:kneser-ney}. In this equation, $c(w_{i-1},w_i)$ represents the count of bi-grams in the training corpus, $\delta$ denotes a discounting factor, $\sum_{w}c(w_{i-1},w)$ signifies the sum of counts of all words $w$ following the word $w_{i-1}$ in the training corpus, and $\lambda_{w-1}$ stands for a normalization factor.

\begin{equation}
p_{KN}(w_{i}|w_{i-1}) = \frac{\max(c(w_{i-1},w_{i}) - \delta, 0)}{\sum_{w'} c(w_{i-1},w')} + \lambda_{w_{i-1}} p_{KN}(w_{i})
\label{eq:kneser-ney}
\end{equation}

In this study, we experimented with both of these types of language modeling. For statistical language models, we used KenLM toolkit \cite{heafield2011kenlm}. KenLM is an n-grams language model toolkit that implements the modified Kneser-Ney smoothing \cite{heafield2013scalable}. The toolkit implements disk-based streaming algorithms allowing it to be 7 times faster than its counterpart SriLM \cite{stolcke2002srilm}, a popular language models toolkit, in estimation with efficient memory management \cite{heafield2013efficient}. We experimented with different orders of grams ranging from bigrams to 6 grams. However, other hyper-parameters are set to defaults. For datasets' splits, each dataset has been split into 90\%, 10\% subsets for training and testing respectively.

Language models are subject to evaluation through two distinct approaches: extrinsic and intrinsic. Extrinsic evaluation involves assessing the model's performance within a downstream task, such as speech recognition or optical character recognition. Conversely, intrinsic evaluation entails self-assessment of the model's competence without reliance on external tasks. Perplexity $\text{PPL}$ serves as a metric for intrinsic evaluation, quantifying the model's surprise when presented with a given sentence, leveraging the probabilities learned during training. The calculation of perplexity is expressed in Equation \ref{eq:perplexity} as follows:

\begin{equation}
\text{PPL}(D, \mathcal{M}) = \exp \left( -\frac{1}{N} \sum_{i=1}^{N} \log \mathcal{M}(w_i) \right)
\label{eq:perplexity}
\end{equation}

Where $\text{PPL}(D, \mathcal{M})$ represents the perplexity of the language model $\mathcal{M}$ on a test dataset $D$, $N$ is the total number of words in the dataset, $w_i$ is the $i$-th word in the dataset, and $\mathcal{M}(w_i)$ is the probability assigned by the language model $\mathcal{M}$ to the $i$-th word.

A lower perplexity value indicates that the language model is better at predicting the given sequence of words.

\subsection{Statistical n-grams}
\label{ssec:ngrams}

In this subsection, we present the results of building language models on dotted and dotless text using an n-grams-based model.

Before discussing the results of the proposed language models, we present our preparation procedure for our datasets. Table \ref{tab:lm_datasets_splits_stats} shows the statistics of these datasets for each split in terms of vocabulary size.

\input{tables/lm_datasets_splits_stats}

It should be noted that we applied some selection criteria for our datasets where some samples were eliminated. For instance, in the poetry dataset, we were interested in studying classical poetry. Hence, we selected samples where the meter is known to be out of the classical poetry meters.

Furthermore, the Wikipedia dataset contains samples with various lengths. To capture meaningful samples, we only selected those with lengths greater than or equal to 30 tokens.

\begin{figure}
    \centering
    \includegraphics[scale=0.45]{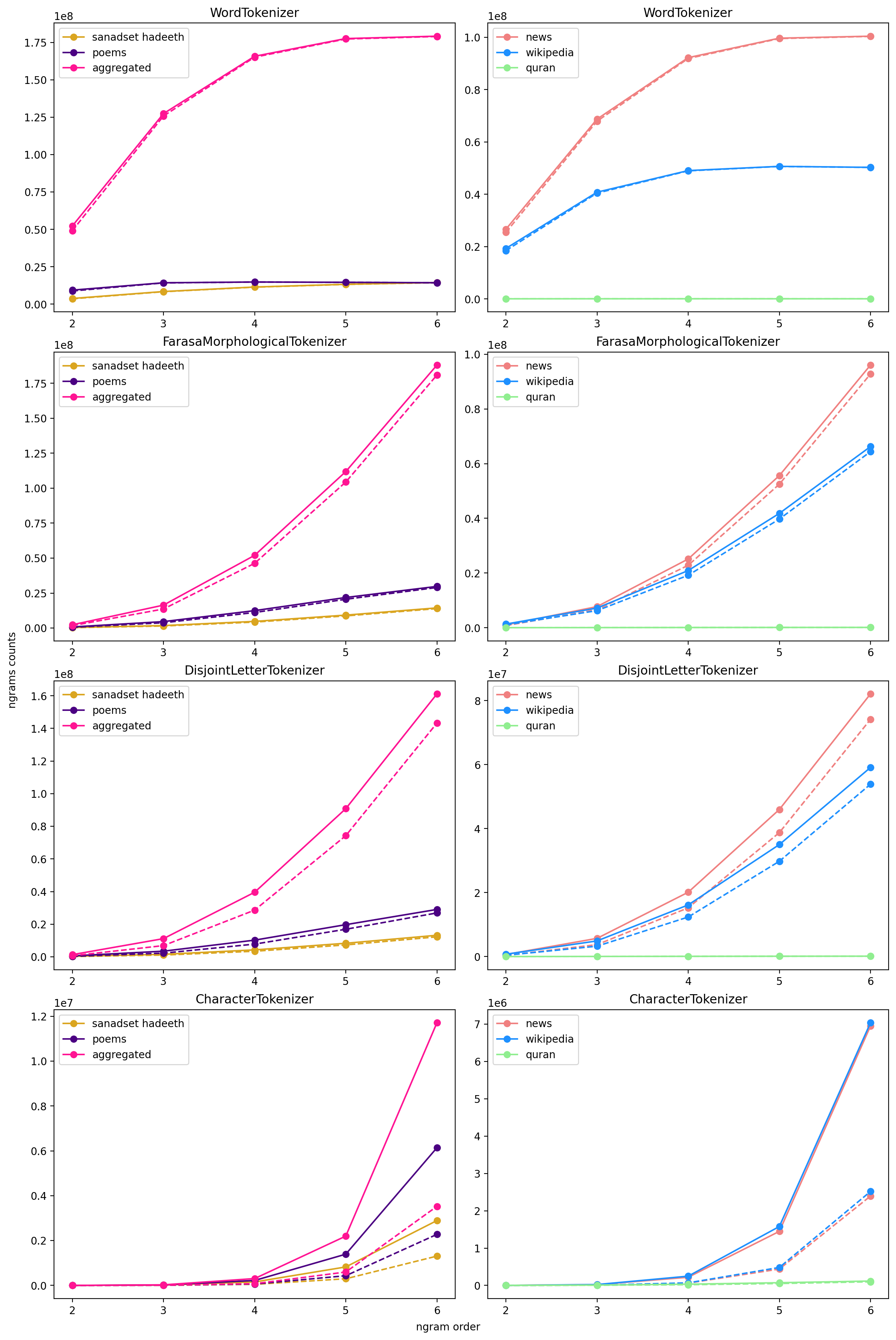}
    \caption{n-grams counts of various tokenizations applied to all proposed datasets}
    \label{fig:ngrams_counts}
\end{figure}

Figure \ref{fig:ngrams_counts} presents the n-grams counts for each order we investigated in this work. In this figure, the dotted line is the n-grams counts for the undotted dataset sharing the same color. Each row in the figure presents the results of a tokenization type with two groups of datasets presented in two plots as it is confusing to plot all in one plot.

A general observation from this graph is that the difference between dotted and dotless counts increases along with the granularity level of the tokenization. That is, as the tokenization is more coarse-grained, the dotless n-grams counts are closer to the dotted ones. 

Another interesting observation is that as the n-grams order grows, the rich datasets' counts tend to decrease faster than the non-rich datasets in both dotted and dotless texts. This can be noticed clearly in poems and Wikipedia datasets as compared to the Sanadset and the news datasets. This behavior is more apparent in coarse-grained tokenization. For instance, the poetry dataset's sixth-order counts are less than its fifth-order. This indicates that although rich datasets have more vocabulary, many of their vocabularies and, subsequently, sequences are rare as compared to non-rich datasets.

Table \ref{tab:ngrams_oovs_stats} presents the Out-Of-Vocabulary (OOVs) statistics of dotted and dotless text after splitting our dataset to training and testing splits for each tokenization scheme. The \textit{ratio} raw presents the dotless to dotted OOVs ratio to measure the reduction in OOVs of dotless text as compared to dotted. It should be noted that the last column represents an aggregated dataset with samples collected from all the datasets we used.

\input{tables/ngrams_oovs_stats}

It can be noticed from this table that the dotless to dotted ratio resembles similar properties for rich datasets as in Tables \ref{tab:words_stats}, \ref{tab:farasa_stats}, and \ref{tab:disjoint_stats}. For instance, It is noticed that the poems dataset has a low ratio across all tokenizations as compared to the Sanadset dataset. Wikipedia, although it has a large running text size as compared to news, has a similar ratio to news ratio. We can also note that farasa tokenization ratios are higher than other tokenizations. The results in this table seem to follow from our conclusions and hypotheses drawn in our discussion in section \ref{sec:text_analysis}.

\begin{figure}
    \centering
    \includegraphics[scale=0.45]{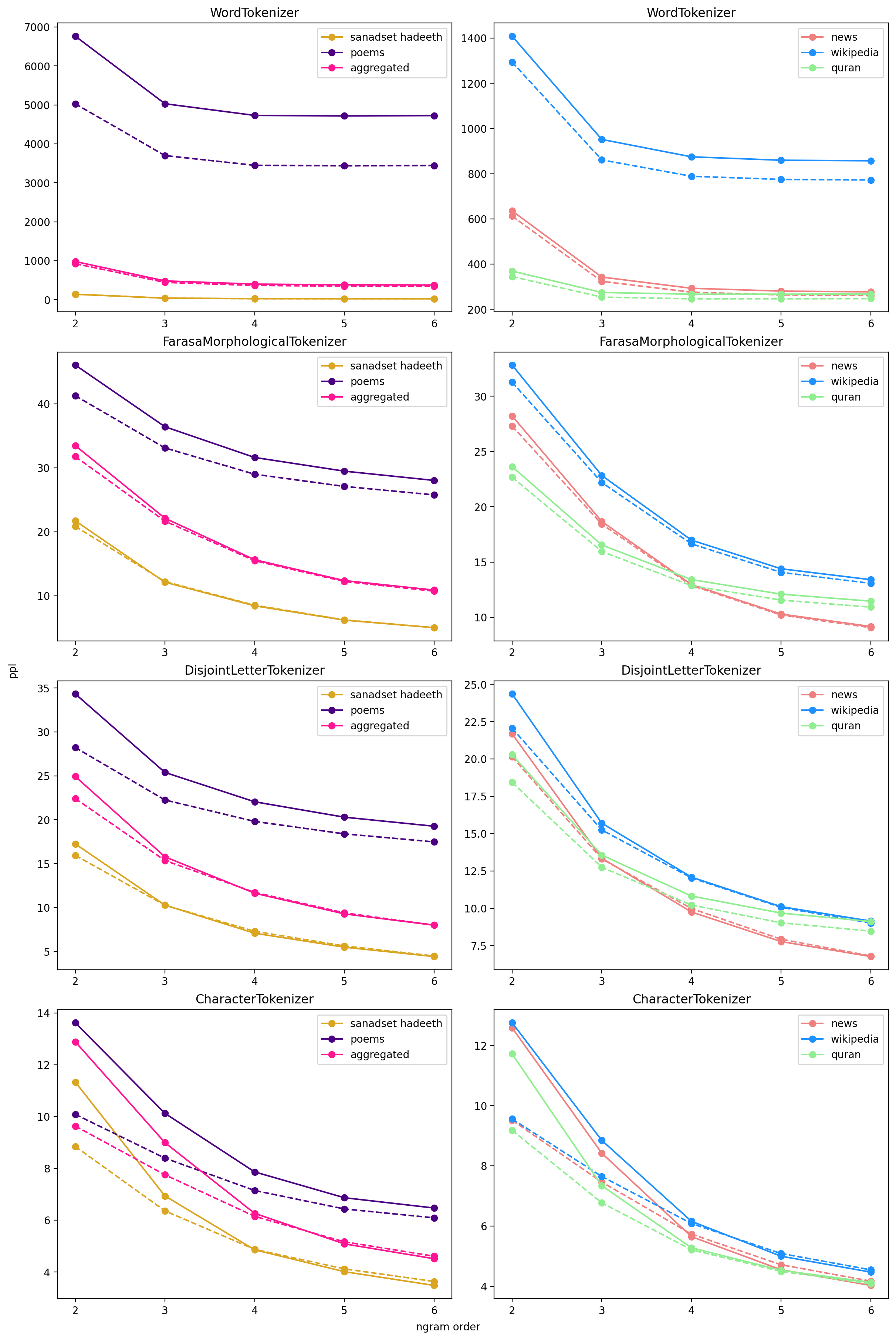}
    \caption{n-grams perplexities of various tokenizations applied to all proposed datasets}
    \label{fig:ngrams_ppl}
\end{figure}

Figure \ref{fig:ngrams_ppl} plots the perplexity of our statistical models trained on all the proposed datasets across different tokenizations with various n-grams orders. As in the n-grams counts figure, the dashed lines represent the dotless version of the dataset.

A general pattern to note in this figure is that as the n-gram order increases, the difference in perplexity between dotted and dotless text decreases. This pattern becomes more apparent as the tokenization is more fine-grained. We can also note that, as the dataset is not rich, i.e. has a smaller V/N, the dotted text perplexity is almost the same as dotless with a higher order of n-grams in almost all tokenizations. This indicates that dotless text may yield the same performance as dotted one in this task.

Unsurprisingly, as the dataset is rich, its perplexity is high. However, it is interesting to note that the perplexity of these rich datasets is notably higher than the aggregated dataset in all tokenizations although they dominate the aggregated dataset as they have more running text N than the non-rich datasets.

\subsection{Neural language models}
\label{ssec:nlm}

\begin{figure}
    \centering
    \includegraphics[scale=0.5]{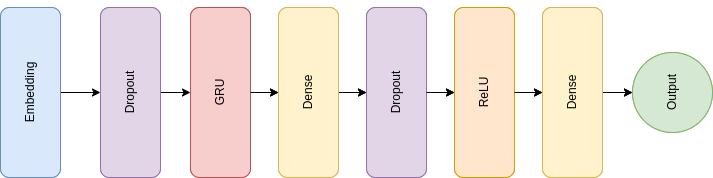}
    \caption{Neural language models architecture}
    \label{fig:nlm_arch}
\end{figure}

For the neural language model, we implemented a GRU-based model. The model comprises an embedding layer, a dropout layer \cite{gal2016theoretically}, a layer of stacked GRU units \cite{chung2014empirical}, a dense layer utilizing ReLU activation, another dropout layer, an additional dense layer, and culminates in an output layer generating the model's log probabilities. We used cross-entropy as our loss function and adam \cite{kingma2014adam} as an optimizer. Figure \ref{fig:nlm_arch} describes this architecture.

As we are experimenting with different datasets from different domains, with different tokenization schemes, the optimal set of hyperparameters may differ from one setup to another. For that reason, we implemented a tuning procedure that determines the best hyperparameters for a given dataset tokenized with a given tokenization scheme. It should be noted that this tuning was applied on the dotted experiment then the resulting best hyperparameters are applied to the dotless experiment. We used AHSH scheduling technique \cite{li2018massively} that terminates bad trials early. We used ray framework \cite{liaw2018tune} which is a well-known framework for distributed computing. We optimized in a grid-search style on a subset of the training set the following hyperparameters:

\begin{itemize}
    \item The number of GRUs chosen from (2, 3, 4).
    \item The GRU hidden size chosen from (256, 512).
    \item The embedding size chosen from (256, 512).
    \item The dropout chosen from (0.2, 0.333, 0.5)
    \item The initial learning rate chosen from (0.01, 0.001).
\end{itemize}

With word tokenization, or more generally any rich vocabulary dataset, the embedding and output layers parameters explode. It has been shown that both of these layers exhibit similar properties \cite{press2016using}, \cite{inan2016tying}, \cite{nowlan1991adaptive},\cite{pappas2018beyond}. Therefore, they can share the same parameters resulting in a noticeable reduction in the model size. This technique is known as weight tying. We used it when applicable based on the best hyperparameters resulting from the implemented tuning procedure. 

We reduced the learning rate on plateaus when little progress has been made. That is, we decrease the learning rate by a factor of 0.75 if no improvement was noticed after a full epoch. The training continued for 100 epochs at most unless no significant improvement with a margin greater than $5\times 10^{-3}$ is witnessed for consecutive 5 epochs. The best model that achieved the lowest validation loss is kept. We built the model with Pytorch \cite{paszke2019pytorch} exploiting the batteries-included features provided by Pytorch-Lightening package \cite{falcon2019pytorch}. For datasets, we partitioned each dataset into 85\%, 5\%, and 10\% splits for training, validation, and testing considering the batch size to be 64.

Vocabulary in the language follows Zipfian distribution \cite{piantadosi2014zipf}, \cite{moreno2016large}. This implies that a large subset of vocabulary has very little frequency while only a small portion of the vocabulary covers most of the running text. In our experiments, we selected vocabulary that covers 95\% of the running text for word tokenization. 

We fixed a sequence length for each dataset with a given tokenization. We discard the remaining text for samples larger than the specified sequence length and padded samples with smaller sequence lengths. This padding token is ignored in loss and perplexity calculations. We noticed that a tiny portion of samples of each sufficiently large dataset is significantly large. The sequence length of the maximum sample will result in lengthy training with little knowledge learned. Additionally, this issue becomes much more apparent with more fine-grained tokenization as GRU learning is less efficient for long sequences \cite{khandelwal2018sharp}, \cite{rumelhart1985learning}, \cite{werbos1990backpropagation}. Hence, we considered different percentiles for different tokenization. We considered the length of the 0.99 percentile sample for word tokenization, 0.975 percentile sample for disjoint letters tokenization, and 0.95 percentile sample for character tokenization. 

Table \ref{tab:nlms_results} presents the results of our neural language models applied on all of the proposed datasets with all proposed tokenizations. In that table, PPL is the perplexity of dotted text while PPL` is the perplexity of the undotted one.
\input{tables/nlms_results}


Analyzing perplexity results from this table, it can be noted that rich datasets have very high perplexity. This can be seen in poems as compared to Sanadset and Wikipedia as compared to news datasets. Poems, as it is the richest dataset has the highest perplexity as compared to any other dataset across all tokenizations. This is because rich datasets are rich in vocabulary and thus are less predictable making it harder for the learning algorithm to learn.



\begin{figure}
    \centering
    \includegraphics[scale=0.35]{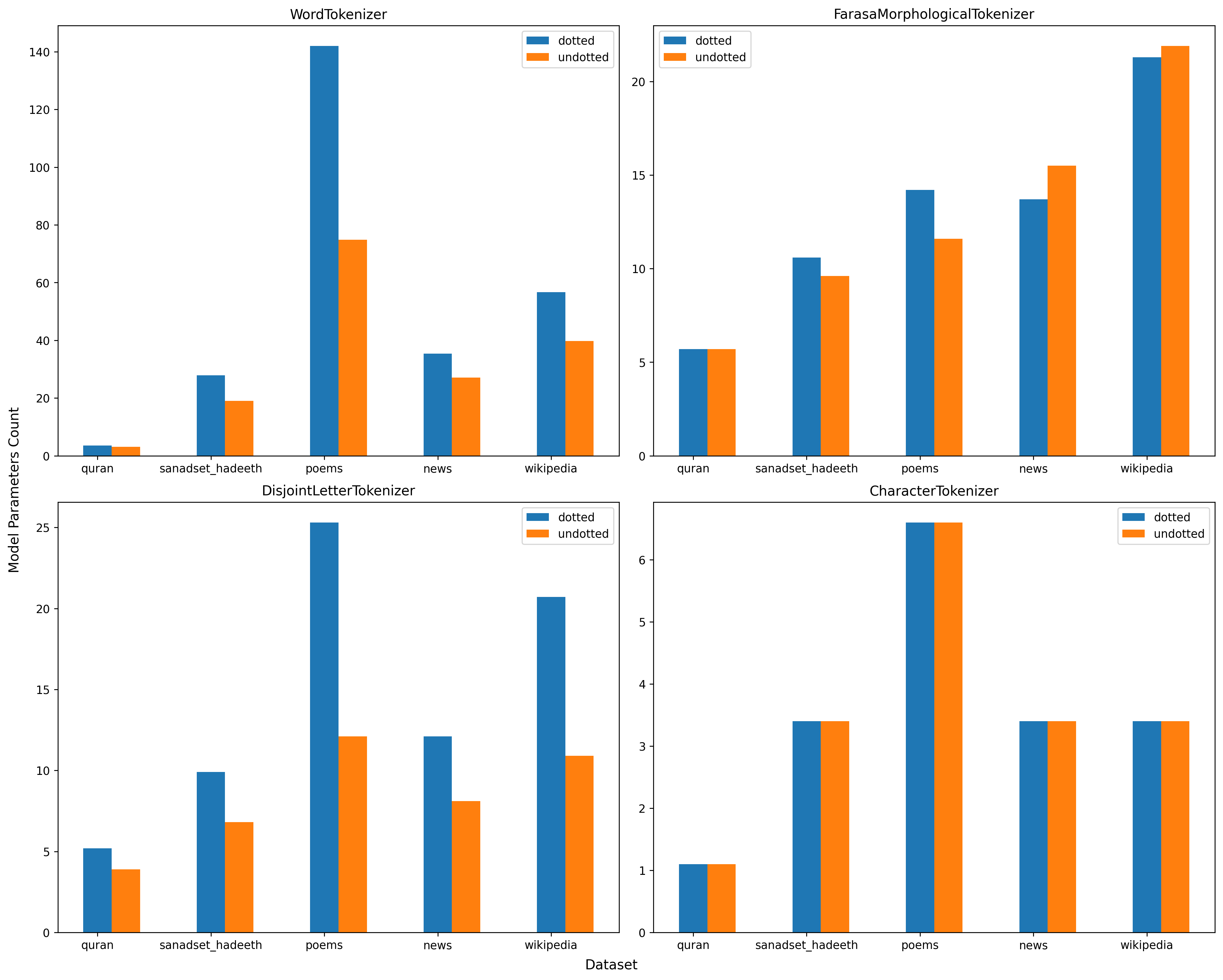}
    \caption{Neural Language Models parameters size comparison for dotted and dotless text}
    \label{fig:nlms_model_size}
\end{figure}

As the dotless text reduces the model vocabulary compared to the dotted, the embedding layer of the dotless text becomes smaller than the dotted embedding. This results in reducing the model parameters.
Figure \ref{fig:nlms_model_size} plots the model parameters size of dotted text as compared to dotless on all our datasets across all tokenizations.

In this figure, it can be noted that the difference in model size is almost negligible for character tokenization. This is because the character's embedding size is generally small as compared to other layers due to the small vocabulary size. Furthermore, the difference between dotless and dotted vocabulary is insignificant.

An interesting observation arises in farasa tokenization, where, in certain datasets, the model size of the dotted representation is unexpectedly smaller than the dotless one. This trend emerges due to the ratio of dotless to dotted vocabulary decreasing as the vocabulary size increases, as illustrated in Figure \ref{fig:dotless_vocab_analysis}. Such findings suggest the significance of carefully selecting the vocabulary size, particularly within the framework of this tokenization method, to effectively manage the dotless and dotted representations.

The figure shows a noticeable reduction in terms of model size, especially for rich datasets as compared to non-rich datasets. This distinction becomes more apparent in coarse-grained tokenizations. Furthermore, the model size difference is negligible for small datasets as compared to mid-size and large-size datasets. This can be clearly seen in quran dataset.

\section{Comparison with English}
\label{sec:compare_with_en}

Arabic, known for its intricate morphology, exhibits a notably rich linguistic structure compared to widely spoken languages like English, resulting in a substantial expansion of the language's vocabulary. As demonstrated previously, the adoption of dotless text yields a reduction in vocabulary, consolidating multiple dotted vocabularies into a singular dotless vocabulary known as \textit{rasm}. To explore and quantify this reduction in vocabularies, we conducted comparative analyses between dotless and dotted text within Arabic, compared against an English corpus. These comparisons were approached in two distinct manners. Firstly, utilizing a parallel corpus, we employed the UN corpus \cite{ziemski2016united} to facilitate this comparison. Secondly, we opted for two corpora drawn from similar domains, selecting the English Wikipedia corpus for comparison with the Arabic corpus. Notably, the English corpus substantially exceeded the Arabic corpus in terms of running text volume. To ensure a fair comparison, we curated a subset of English articles whose cumulative text size closely approximated that of the Arabic corpus. In both comparisons, the English text underwent processing by converting characters to lowercase and eliminating characters not belonging to the English character set, including numerals and punctuation. The statistics detailing these datasets are presented in Table \ref{tab:parallel_corpus}.

\input{tables/parallel_corpus}

From this table, it can be seen that English has a lower character entropy than dotted Arabic, but dotless is the lowest. However, dotless Arabic has a higher entropy in words vocabulary very close to Wikipedia. English here is the lowest. This indicates that English generally is more predictable than dotted and dotless Arabic for word vocabulary. It is also interesting to note that although the UN corpus is parallel, English running text is 40M more, around 17\%, in terms of running text, N.

\begin{figure}[!ht]

\begin{subfigure}{.475\linewidth}
  \includegraphics[width=\linewidth]{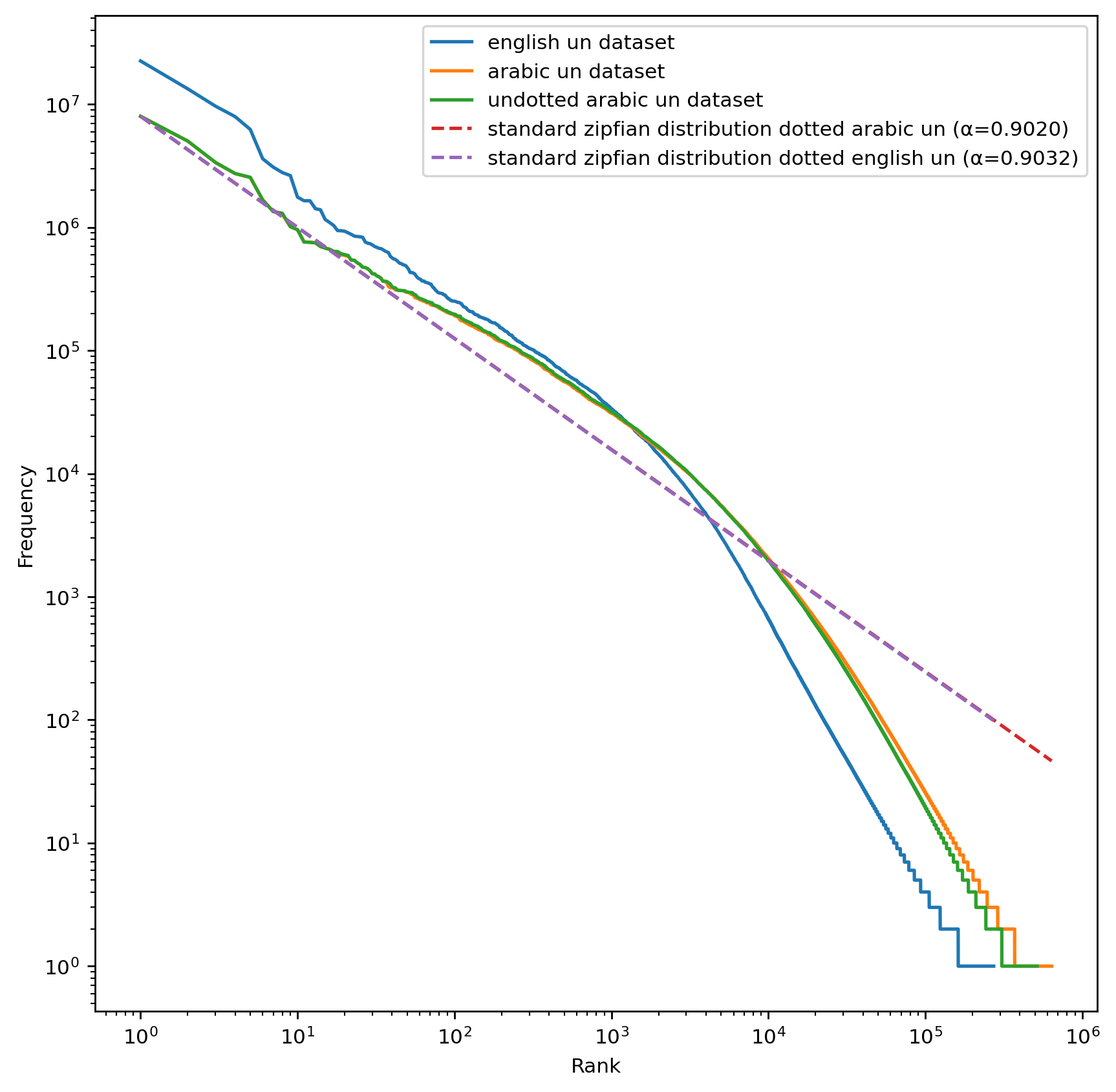}
  \caption{UN parallel Zipf's law}
  \label{fig:un_parallel_zipf}
\end{subfigure}\hfill 
\begin{subfigure}{.475\linewidth}
  \includegraphics[width=\linewidth]{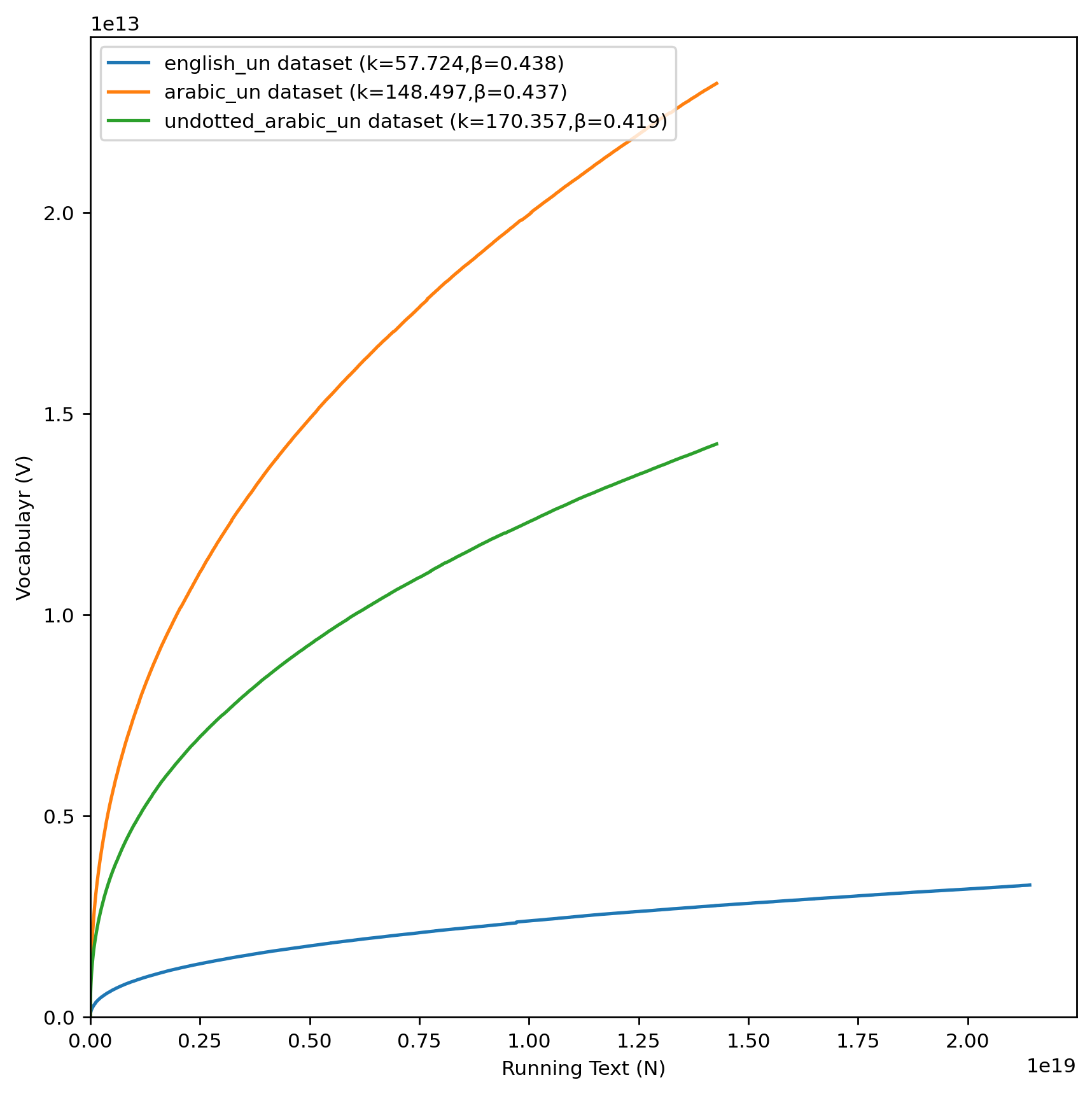}
  \caption{UN parallel Heap's law}
  \label{fig:un_parallel_heap}
\end{subfigure}

\medskip 
\begin{subfigure}{.475\linewidth}
  \includegraphics[width=\linewidth]{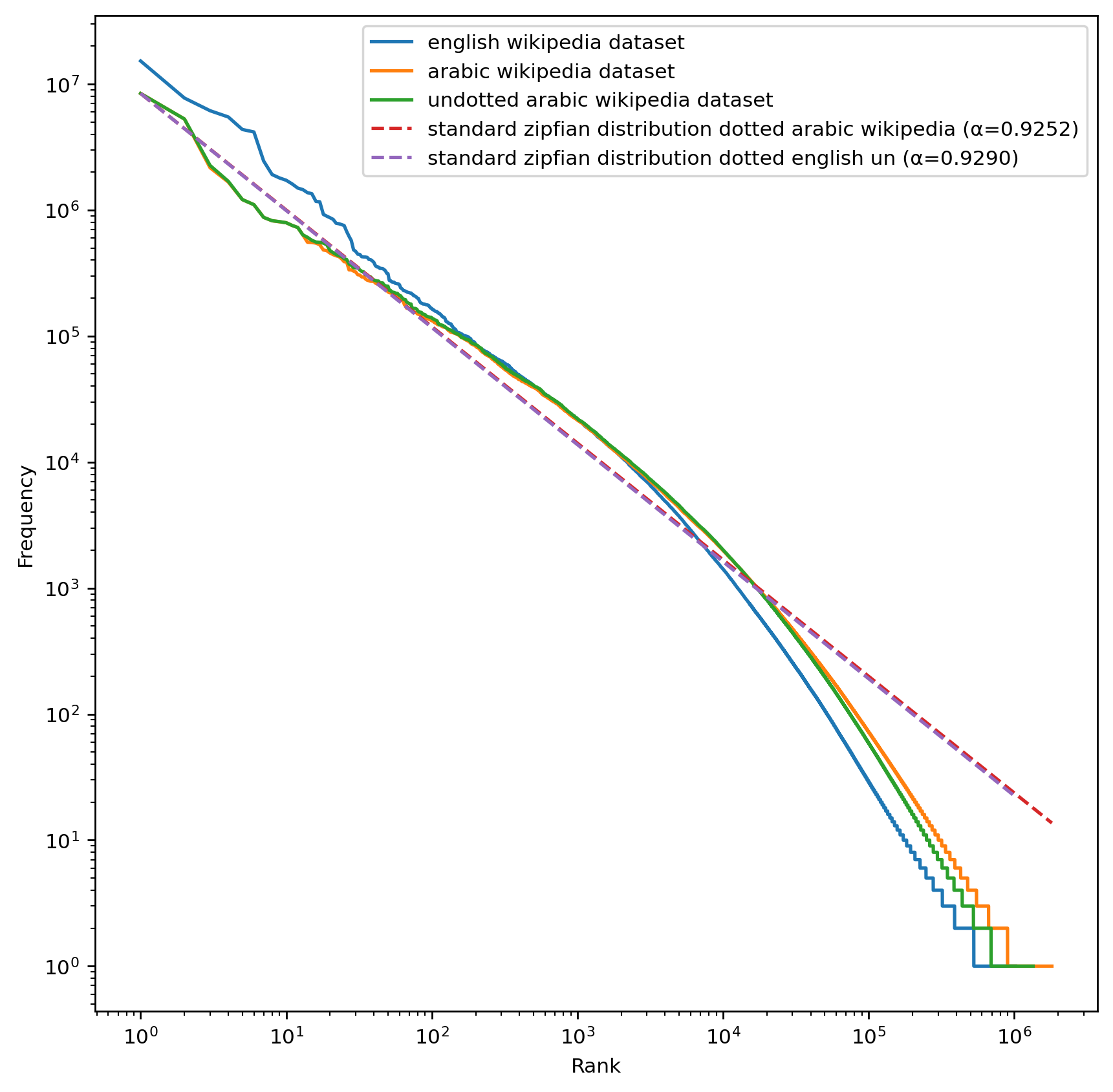}
  \caption{Wikipedia Zipf's law}
  \label{fig:wikipedia_parallel_zipf}
\end{subfigure}\hfill 
\begin{subfigure}{.475\linewidth}
  \includegraphics[width=\linewidth]{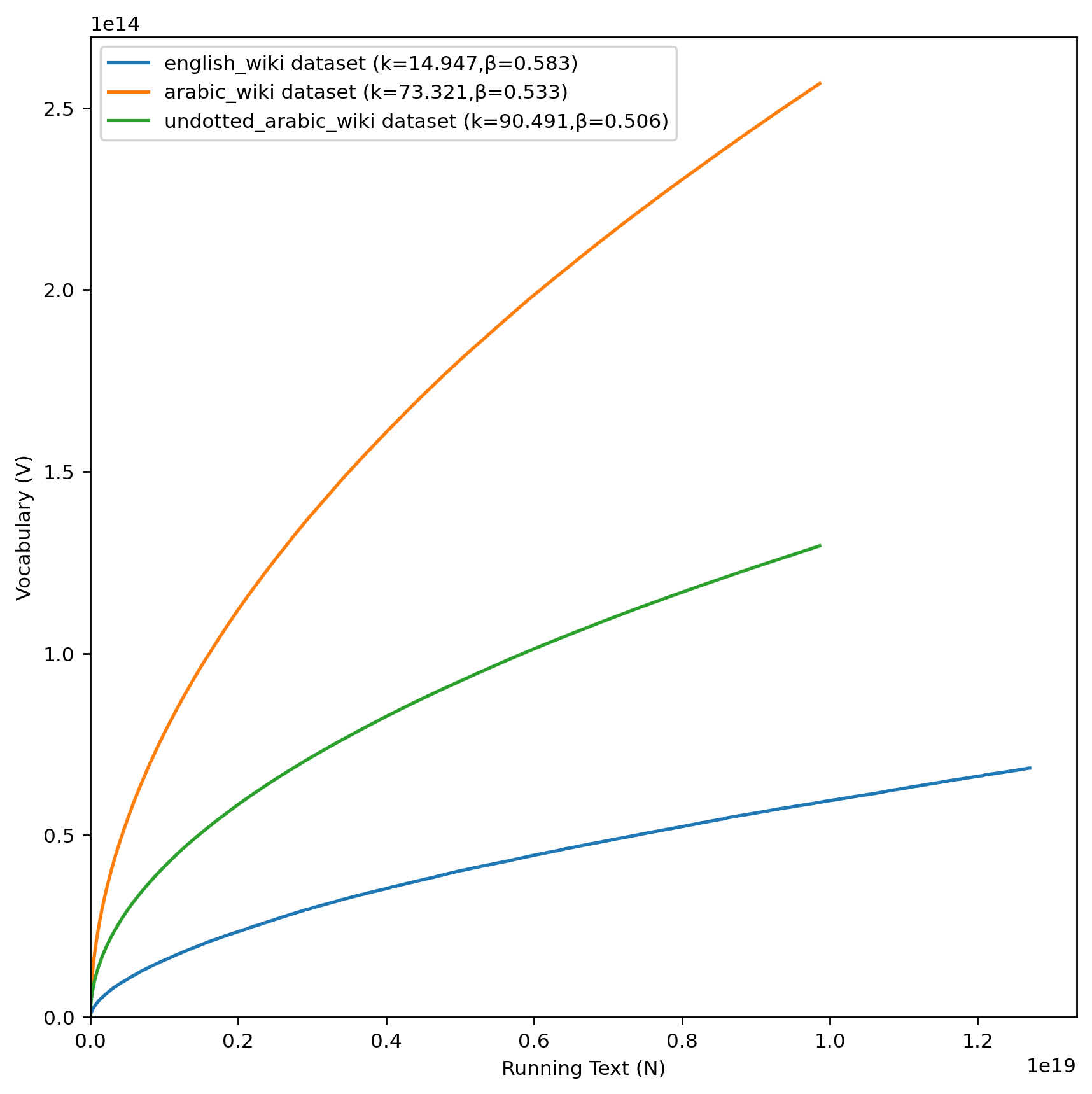}
  \caption{Wikipedia Heap's law}
  \label{fig:wikipedia_parallel_heap}
\end{subfigure}

\caption{Zipf's and Heap's law for English-Arabic Parallel datasets}
\label{fig:parallel_datasets_zipf_heap}
\end{figure}

To further study vocabulary frequencies and growth, we plot Zipf's and Heap's law for both of the parallel datasets in figure \ref{fig:parallel_datasets_zipf_heap}. From this graph, we can notice that the Zipfian fit is overlapping for both datasets. It is interesting to generalize this across languages. It can be noted that in UN corpus, the English graph seems to be shifted upward for the most frequent vocabulary and downward for the least frequent vocabulary. That is, the graphs for both languages are identical if the shift effect is neglected. Following a similar pattern for Wikipedia dataset, it seems that the most frequent vocabulary for English is more frequent than Arabic. However, the least frequent English vocabularies are less frequent than Arabic.




For the vocabulary growth described by Heap's law, it can be seen that Arabic is larger by order of magnitudes in terms of vocabulary although the English running text is larger for both datasets, UN and Wikipedia. It can also be noted from the graph that the Wikipedia dataset is richer than the UN dataset. Following our previous findings that the vocabulary growth for dotless text is slower for rich datasets, we can see this pattern clearly in the UN dataset as compared to Wikipedia. In fact, in Wikipedia, the dotless text growth is close to the average of both English and Arabic vocabulary growth.



\section{Conclusion and future directions}
\label{sec:conclusion}

In this research paper, we introduce an alternative representation of Arabic text for NLP, rooted from the idea of removing dots. Our investigation involves a comprehensive comparison between this new dotless representation and the conventional dotted Arabic text across five diverse datasets encompassing different domains and sizes, employing various tokenization schemes. As a case study within the realm of NLP tasks, we conduct a comparative analysis between dotted and dotless text specifically focused on language modeling. We performed the experimentation with two different language modeling techniques with various datasets and tokenizations. Our findings, briefly summarized below, have been extensively discussed in detail earlier in the paper.

Analyzing token statistics reveals that the ratio of dotless vocabulary to dotted varies depending on the utilized tokenization method. When considering the aggregated dataset as a representation of the language, the observed ratio of V` to V amounts to 68.1\%, indicating a reduction of approximately 22\% in vocabulary attributed to the removal of dots. Notably, this reduction tends to increase as the dataset richness grows. Moreover, a similar reduction pattern is observed across various levels, specifically for farasa and disjoint letters tokenization. Among these, the disjoint letters tokenization yields the most significant vocabulary reduction, reaching approximately 60\% on the aggregated dataset.


All types of tokenizations presented in this work follow Zipf's law. The more coarse-grained the tokenization, the more Zipfian it is. Dotless vocabulary frequencies have similar behavior to dotted ones in Zipfian settings. This observation is dataset and tokenization-independent. Dotted and dotless vocabulary adheres to heap's law. However, dotless growth is less than dotted. This is more apparent for rich datasets. Further, disjoint growth is much slower confirming our previous findings of V`/V being the smallest.

The undotted text holds entropy close to the dotted text with a maximum difference of less than 10\% in character tokenization. However, it reduces the character set by a margin of 61.3\%. Based on the entropy analysis of dotted text as compared to dotted text, the dots reduced text redundancy by 5\%. This implies that there are other sources of redundancy other than dots that are also present in the dotless text. The dotted text has a slightly higher entropy compared to the dotless text. This is dataset size and tokenization independent. Also, this applies to rich and non-rich datasets.

Although dotless characters' entropy is the lowest, English words' entropy is almost one-third lower than dotted and dotless Arabic words' entropy. For a parallel English-Arabic corpus, English's most frequent vocabularies are more than the Arabic most frequent ones and English's least frequent vocabularies are less than those of Arabic. This implies, the vocabulary difference is not uniform.



The difference between dotless and dotted n-grams increases as the tokenization becomes fine-grained. This supports our previous hypothesis that fine-grained tokenization covers more dotted text as compared to coarse-grained. However, this difference is generally small.Moreover, as the n-grams order grows, rich datasets’ n-grams counts decrease faster than the non-rich datasets. This applies to dotted and dotless text.

Generally, for statistical language models, the dotless text perplexity becomes closer to dotted as the n-grams order grows. This observation becomes clearer as the tokenization becomes more fine-grained. Characters' perplexity for neural language models is very close for dotted and dotless text, closer than any other tokenization.

In neural language models, dotless text consistently reduces the model size due to the reduction in vocabulary across various tokenization methods, with one notable exception found in farasa tokenization. In specific cases within farasa tokenization, the dotless vocabulary does not exhibit a reduction compared to the dotted version, due to the choice of vocabulary size. Moreover, the impact of dotless vocabulary is minimal in character tokenization, as the variance in vocabulary size between dotless and dotted representations is negligible in this tokenization scheme.


As a future work, it is interesting to study the performance of the dotless text on downstream tasks such as text classification, POS tagging and machine translation. It is also interesting to extend the use of this representation to other tasks like speech recognition and optical character recognition and compare the performance with standard text representation including dots.

\section*{Acknowledgment}
The authors would like to thank Saudi Data and AI Authority (SDAIA) and King Fahd University of Petroleum and Minerals (KFUPM) for supporting this work through SDAIA-KFUPM Joint Research Center for Artificial Intelligence grant number JRC--AI--RFP--10.

\noindent
\starttwocolumn
\bibliography{compling_style}

\end{document}

%% file: tables/datasets_statistics.tex

\begin{table}[]
\centering
\begin{tabular}{lllllcc}
\hline
\textbf{Dataset} & \textbf{$N_w$} & \textbf{$V_w$} & \textbf{$V`_w$} & \textbf{$N_c$} & \multicolumn{1}{l}{\textbf{$V_c$}} & \multicolumn{1}{l}{\textbf{$V`_c$}} \\ \hline
Quran            & 77,797         & 14,748         & 13,229          & 330,709        & \multirow{6}{*}{31}                & \multirow{6}{*}{19}                 \\ \cline{1-5}
Sanadset         & 28,880,818     & 317,331        & 227,590         & 112,879,152    &                                    &                                     \\ \cline{1-5}
Poems            & 34,940,290     & 1,007,279      & 631,487         & 145,574,240    &                                    &                                     \\ \cline{1-5}
News             & 134,862,816    & 892,583        & 654,982         & 663,796,116    &                                    &                                     \\ \cline{1-5}
Wikipedia        & 177,422,512    & 1,811,244      & 1,345,853       & 851,699,872    &                                    &                                     \\ \cline{1-5}
Aggregated       & 376,184,233    & 2,739,172      & 1,865,126       & 1,774,280,089  &                                    &                                     \\ \hline
\end{tabular}
\caption{Datasets vocabulary and characters statistics, dotted and dotless}
\label{tab:datastes_stats}
\end{table}

%% file: tables/undotted_mapping_table.tex
\begin{table}
\centering
\begin{tabular}{|c|c|c|c|c|c|c|c|}
\hline
\textbf{Letter} & Mapping                 & \textbf{Letter} & Mapping                 & \textbf{Letter} & Mapping                 & \textbf{Letter} & Mapping \\ \hline
\RL{ا}          & \RL{ا}                  & \RL{د}          & \multirow{2}{*}{\RL{د}} & \RL{ض}          & \RL{ص}                  & \RL{ك}          & \RL{ك}  \\ \cline{1-3} \cline{5-8} 
\RL{ب}          & \multirow{3}{*}{\RL{ٮ}} & \RL{ذ}          &                         & \RL{ط}          & \multirow{2}{*}{\RL{ط}} & \RL{ل}          & \RL{ل}  \\ \cline{1-1} \cline{3-5} \cline{7-8} 
\RL{ت}          &                         & \RL{ر}          & \multirow{2}{*}{\RL{ر}} & \RL{ظ}          &                         & \RL{م}          & \RL{م}  \\ \cline{1-1} \cline{3-3} \cline{5-8} 
\RL{ث}          &                         & \RL{ز}          &                         & \RL{ع}          & \multirow{2}{*}{\RL{ع}} & \RL{ن}          & \RL{ں}  \\ \cline{1-5} \cline{7-8} 
\RL{ج}          & \multirow{3}{*}{\RL{ح}} & \RL{س}          & \multirow{2}{*}{\RL{س}} & \RL{غ}          &                         & \RL{ه}          & \RL{ه}  \\ \cline{1-1} \cline{3-3} \cline{5-8} 
\RL{ح}          &                         & \RL{ش}          &                         & \RL{ف}          & \RL{ڡ}                  & \RL{و}          & \RL{و}  \\ \cline{1-1} \cline{3-8} 
\RL{خ}          &                         & \RL{ص}          & \RL{ص}                  & \RL{ق}          & \RL{ٯ}                  & \RL{ي}          & \RL{ى}  \\ \hline
\end{tabular}
\caption{Arabic dotted letters mapped to their dotless variant}
\label{tab:undotting_mapping}
\end{table}

%% file: tables/tokens_stats/words.tex
\begin{table}[h]
\centering
\begin{tabular}{cllllllll}
\toprule
\textbf{Dataset}   & \textbf{V} & \textbf{N} & \textbf{V/N} & \textbf{V`} & \textbf{V`/N} & \textbf{V`/V (\%)} & \textbf{S} & \textbf{S`} \\ 
\midrule
\textbf{Quran}      & 14,748     & 77,797     & 18.96        & 13,229      & 17            & 89.70               & 5.39       & 5.49        \\ 
\textbf{Sanadset}   & 317,331    & 28.88M     & 1.10         & 227,590     & 0.79          & 71.72               & 5.95       & 6.23        \\ 
\textbf{Ashaar}     & 1,007,279  & 34.94M     & 2.88         & 631,487     & 1.81          & 62.69               & 6.13       & 6.49        \\ 
\textbf{Wikipedia}  & 1,811,244  & 177.42M    & 1.02         & 1,345,853   & 0.76          & 74.31               & 7.33       & 7.82        \\ 
\textbf{News}       & 892,583    & 134.86M    & 0.66         & 654,982     & 0.49          & 73.38               & 6.73       & 7.09        \\ 
\textbf{Aggregated} & 2,739,172  & 376.18M    & 0.73         & 1,865,126   & 0.5           & 68.09               & 7.12       & 7.66        \\
\bottomrule
\end{tabular}
\caption{Datasets statistics with words tokenization covering dotted and dotless vocabulary}
\label{tab:words_stats}
\end{table}

%% file: tables/tokens_stats/farasa.tex
\begin{table}[h]
\centering
\begin{tabular}{cllllllll}
\toprule
\textbf{Dataset}    & \textbf{V} & \textbf{N}  & \textbf{V/N} & \textbf{V`} & \textbf{V`/N} & \textbf{V`/V (\%)} & \textbf{S} & \textbf{S`} \\
\midrule
\textbf{Quran}      & 7,677      & 129,788     & 5.92         & 6,074       & 4.68          & 79.04              & 4.46       & 4.65        \\ 
\textbf{Sanadset}   & 105,469    & 44,069,593  & 2.39         & 74,674      & 0.17          & 70.85              & 5.50       & 5.89        \\ 
\textbf{Ashaar}     & 361,739    & 57,350,212  & 0.63         & 242,735     & 0.42          & 67.09              & 5.79       & 6.22        \\ 
\textbf{Wikipedia}  & 1,002,148  & 303,831,223 & 0.33         & 791,084     & 0.26          & 78.91              & 7.32       & 7.81        \\ 
\textbf{News}       & 344,627    & 241,222,485 & 0.14         & 263,143     & 0.11          & 76.36              & 6.45       & 6.88        \\ 
\textbf{Aggregated} & 1,417,437  & 646,603,301 & 0.22         & 1,061,481   & 0.16          & 74.92              & 7.14       & 7.66        \\
\bottomrule
\end{tabular}
\caption{Datasets statistics with farasa tokenization covering dotted and dotless vocabulary}
\label{tab:farasa_stats}
\end{table}

%% file: tables/tokens_stats/disjoint.tex
\begin{table}[h]
\centering
\begin{tabular}{cllllllll}
\toprule
\textbf{Dataset}    & \textbf{V} & \textbf{N}  & \textbf{V/N} & \textbf{V'} & \textbf{V'/N} & \textbf{V`/V (\%)} & \textbf{S} & \textbf{S`} \\
\midrule
\textbf{Quran}      & 6,484      & 164,609     & 3.94         & 4,303       & 2.61          & 66.36              & 4.06       & 4.29        \\
\textbf{Sanadset}   & 76,495     & 55,892,178  & 0.14         & 37,412      & 0.07          & 48.91              & 4.95       & 5.31        \\
\textbf{Ashaar}     & 191,393    & 72,437,612  & 0.26         & 79,968      & 0.11          & 41.78              & 5.16       & 5.56        \\
\textbf{Wikipedia}  & 266,822    & 412,843,347 & 0.06         & 118,150     & 0.03          & 44.28              & 5.54       & 6.07        \\
\textbf{News}       & 151,557    & 326,024,821 & 0.05         & 68,943      & 0.02          & 45.49              & 5.31       & 5.75        \\
\textbf{Aggregated} & 402,019    & 867,375,567 & 0.05         & 162,389     & 0.02          & 40.39              & 5.59       & 6.11        \\
\bottomrule
\end{tabular}
\caption{Datasets statistics with disjoint-letters tokenization covering dotted and dotless vocabulary}
\label{tab:disjoint_stats}
\end{table}

%% file: tables/tokens_stats/entropy.tex
\begin{table}[]
\centering
\begin{tabular}{ccccccccc}
\toprule
\multirow{2}{*}{\textbf{Dataset}} & \multicolumn{2}{c}{\textbf{Words}} & \multicolumn{2}{c}{\textbf{Characters}} & \multicolumn{2}{c}{\textbf{Farasa}} & \multicolumn{2}{c}{\textbf{Disjoint}} \\ \cline{2-9} 
                                  & H                & H`              & H                  & H`                 & H                & H`               & H                 & H`                \\
\midrule
\textbf{Quran}                             & 11.02            & 10.87           & 4.15               & 3.83               & 8.240            & 7.968            & 7.451             & 6.958             \\
\textbf{Sanadset}                          & 10.92            & 10.67           & 4.24               & 3.86               & 8.364            & 8.004            & 7.644             & 7.098             \\
\textbf{Ashaar}                            & 14.33            & 13.70           & 4.3                & 3.9                & 9.659            & 9.148            & 8.480             & 7.687             \\
\textbf{Wikipedia}                         & 13.20            & 12.94           & 4.27               & 3.87               & 8.892            & 8.599            & 8.153             & 7.493             \\
\textbf{News}                              & 13.10            & 12.87           & 4.25               & 3.85               & 8.563            & 8.286            & 7.981             & 7.363             \\
\textbf{Aggregated}                        & 13.6             & 13.26           & 4.27               & 3.87               & 9.063            & 8.711            & 8.210             & 7.523             \\
\bottomrule
\end{tabular}
\caption{Entropy results for each tokenization across the proposed datasets}
\label{tab:entropy_stats}
\end{table}

%% file: tables/lm_datasets_splits_stats.tex
\begin{table}[]
\centering
\begin{tabular}{lllll}
\hline
\textbf{Dataset}                    & \textbf{Text Type} & \textbf{Train Vocab} & \textbf{Val Vocab} & \textbf{Test Vocab} \\ \hline
\multirow{2}{*}{\textbf{Quran}}     & Dotted             & 13,435               & 1,720              & 3,161               \\ \cline{2-5} 
                                    & Dotless            & 12,102               & 1,658              & 3,018               \\ \hline
\multirow{2}{*}{\textbf{Sanadset}}  & Dotted             & 299,162              & 81,763             & 121,488             \\ \cline{2-5} 
                                    & Dotless            & 215,758              & 65,941             & 95,132              \\ \hline
\multirow{2}{*}{\textbf{Poems}}     & Dotted             & 881,669              & 208,236            & 321,650             \\ \cline{2-5} 
                                    & Dotless            & 560,069              & 151,916            & 225,293             \\ \hline
\multirow{2}{*}{\textbf{Wikipedia}} & Dotted             & 970,567              & 211,025            & 323,331             \\ \cline{2-5} 
                                    & Dotless            & 730,885              & 173,672            & 259,353             \\ \hline
\multirow{2}{*}{\textbf{News}}      & Dotted             & 837,244              & 231,783            & 333,842             \\ \cline{2-5} 
                                    & Dotless            & 617,593              & 187,819            & 263,524             \\ \hline
\end{tabular}
\caption{Train,validation, and test statistics for datasets used in language modeling tasks}
\label{tab:lm_datasets_splits_stats}
\end{table}

%% file: tables/ngrams_oovs_stats.tex
\begin{table}[!ht]
\centering
\begin{tabular}{cc|cccccc}
\multicolumn{2}{c|}{\textbf{Tokenization}}                       & \textbf{quran} & \textbf{sanadset} & \textbf{poems} & \textbf{wikipedia} & \textbf{news} & \textbf{aggregated} \\ \hline
\multirow{3}{*}{\textbf{Words}}    & \textit{\textbf{Dotted}}    & 910            & 13,273            & 34,577         & 57,329             & 39,177        & 94,216              \\ \cline{2-8} 
                                   & \textit{\textbf{Dotless}}   & 786            & 8,619             & 20,080         & 40,901             & 26,507         & 60,093              \\ \cline{2-8} 
                                   & \textit{\textbf{Ratio(\%)}} & 86.37          & 64.94             & 58.07          & 71.34              & 67.66         & 63.78               \\ \hline \hline
\multirow{3}{*}{\textbf{Farasa}}   & \textit{\textbf{Dotted}}    & 390            & 4,253             & 12,514         & 32,153             & 17,997        & 51,268              \\ \cline{2-8} 
                                   & \textit{\textbf{Dotless}}   & 300            & 2,948             & 8,112         & 25,104             & 13,680        & 37,430              \\ \cline{2-8} 
                                   & \textit{\textbf{Ratio(\%)}} & 76.92          & 69.32             & 64.82          & 78.08              & 76.01         & 73.01               \\ \hline \hline
\multirow{3}{*}{\textbf{Disjoint}} & \textit{\textbf{Dotted}}    & 287            & 2,534             & 5,985          & 7,747              & 5,619         & 12,196              \\ \cline{2-8} 
                                   & \textit{\textbf{Dotless}}   & 185            & 1,078             & 2,216          & 3,116              & 2,252         & 4,417               \\ \cline{2-8} 
                                   & \textit{\textbf{Ratio(\%)}} & 64.46          & 42.54             & 37.03          & 40.22              & 40.08         & 36.22               \\ \hline \hline
\end{tabular}
\caption{OOVs statistics with dotted and dotless texts across different tokenizations and datasets}
\label{tab:ngrams_oovs_stats}
\end{table}

%% file: tables/nlms_results.tex
\begin{table}[]
\centering
\begin{tabular}{cccccc}
\hline
\multicolumn{2}{c}{Dataset}       & Words   & Farasa & Disjoint & Characters \\ \hline
\multirow{2}{*}{Quran}     & PPL  & 179.02  & 22.89  & 15.76    & 5.28       \\ \cline{2-6} 
                           & PPL` & 195.69  & 18.54  & 14.90    & 4.66       \\ \hline
\multirow{2}{*}{Sanadset}  & PPL  & 33.11   & 5.16   & 4.87     & 2.79       \\ \cline{2-6} 
                           & PPL` & 35.85   & 5.31   & 5.11     & 2.74       \\ \hline
\multirow{2}{*}{Poems}     & PPL  & 1195.47 & 36.60  & 18.30    & 6.17       \\ \cline{2-6} 
                           & PPL` & 1022.35 & 31.4   & 16.8     & 5.6        \\ \hline
\multirow{2}{*}{News}      & PPL  & 166.90  & 7.72   & 7        & 3.17       \\ \cline{2-6} 
                           & PPL` & 179.09  & 8.6    & 7.29     & 3.18       \\ \hline
\multirow{2}{*}{Wikipedia} & PPL  & 290.25  & 9.55   & 8.79     & 3.61       \\ \cline{2-6} 
                           & PPL` & 296.16  & 10.49  & 9.13     & 3.52       \\ \hline
\end{tabular}
\caption{Neural Language Models results of all datasets with different tokenizations}
\label{tab:nlms_results}
\end{table}

%% file: tables/parallel_corpus.tex
\begin{table}[!ht]
\centering
\begin{tabular}{ccccc}
\textbf{Dataset}                  & \textbf{Char Entropy} & \textbf{V}         & \textbf{N}                            & \textbf{Words Entropy} \\ \hline \hline
English UN               & 4.12         & 272,567   & 248,442,788                  & 9.7           \\ \hline
Arabic UN                & 4.19         & 637,778   & \multirow{2}{*}{208,325,620} & 12.21         \\ \cline{1-3} \cline{5-5} 
Dotless Arabic UN        & 3.68         & 515,946   &                              & 12.07         \\ \hline \hline
English Wikipedia        & 4.17         & 1,019,960 & 198,029,572                  & 10.97         \\ \hline
Arabic Wikipedia         & 4.27         & 1,811,244 & \multirow{2}{*}{177,422,512} & 13.2          \\ \cline{1-3} \cline{5-5} 
Dotless Arabic Wikipedia & 3.72          & 1,345,853 &                              & 12.934         \\ \hline
\end{tabular}
\caption{Parallel Datasets Comparison considering dotted and dotless text}
\label{tab:parallel_corpus}
\end{table}